%% file: main.tex
\documentclass{article}
\usepackage{iclr2026_conference,times}
\input{math_commands.tex}

\usepackage{amsmath,amssymb,amsthm,mathtools,bm}
\usepackage{graphicx,booktabs,multirow,array}
\usepackage{algorithm,algorithmic}
\usepackage{microtype,xspace,enumitem}
\usepackage{hyperref,url}
\usepackage{natbib}

\theoremstyle{plain}
\newtheorem{aactheorem}{Theorem}[section]
\newtheorem{aacproposition}[aactheorem]{Proposition}
\newtheorem{aaclemma}[aactheorem]{Lemma}
\newtheorem{aaccorollary}[aactheorem]{Corollary}
\theoremstyle{remark}

\providecommand{\E}{\mathbb{E}}
\providecommand{\KL}{D_{\mathrm{KL}}}
\providecommand{\Unif}{\operatorname{Unif}}

\title{Advanced Policies: A First-Principles Path from Policy Gradient to Q-Learning}
\author{Donghoon Lee}
\iclrfinalcopy
\hypersetup{pdfauthor={Donghoon Lee},
  pdftitle={Advanced Policies: A First-Principles Path from Policy Gradient to Q-Learning}}

\begin{document}
\maketitle
\fancyhead{}

\input{sections/abstract}
\input{sections/revision_note}
\input{sections/introduction}
\input{sections/related_work}
\input{sections/preliminaries}
\input{sections/entropy_augmented_rl}
\input{sections/advanced_policy}
\input{sections/algorithm}
\input{sections/experiments}
\input{sections/conclusion}
\input{sections/ai_assistance}
\input{sections/reproducibility}

\bibliography{iclr2026_conference}
\bibliographystyle{iclr2026_conference}

\appendix
\input{sections/appendix}

\end{document}

%% file: math_commands.tex
\usepackage{amsmath,amsfonts,bm}

\def\1{\bm{1}}

\DeclareMathAlphabet{\mathsfit}{\encodingdefault}{\sfdefault}{m}{sl}
\SetMathAlphabet{\mathsfit}{bold}{\encodingdefault}{\sfdefault}{bx}{n}

\newcommand{\E}{\mathbb{E}}

\newcommand{\softmax}{\mathrm{softmax}}

\newcommand{\KL}{D_{\mathrm{KL}}}
\newcommand{\Var}{\mathrm{Var}}



%% file: sections/abstract.tex
\begin{abstract}
We apply entropy to reinforcement learning in two ways: entropy-augmented reward
defines the soft objective, while relative-entropy regularization
controls the size of a policy-improvement step without changing that
objective. Together they yield a one-parameter family of \emph{advanced
policies} connecting a base policy to its soft-greedy policy. Under
exact evaluation, every nontrivial member improves upon the common base,
although the improvement need not be monotone along the family. Under policy--value 
consistency, this family forms an entropic mirror-descent path. We
then relax this consistency, treating the policy and action-value function 
as independent coordinates of the advanced policy. 
Differentiating the objective of the advanced policy yields 
Advanced Actor--Critic (AAC), whose endpoint gradients recover soft policy
gradient and an action-centered Q-learning-like update. We develop
implementations for discrete and continuous actions. Experiments 
demonstrate effective learning at both endpoints and intermediate parameter values.
\end{abstract}

%% file: sections/revision_note.tex
\begin{quote}
\small
\textbf{Revision note.}
This version substantially revises
\href{https://arxiv.org/abs/2005.08844v2}{arXiv:2005.08844v2}.
It corrects the earlier claim of monotonic return improvement along the
advanced-policy family (Theorem~3) and the associated simultaneous-optimality
claim (Corollary~1). Under exact policy evaluation, the revised analysis
establishes improvement relative to a common base policy and a common
soft-optimal fixed point for the nontrivial improvement operators, without
asserting that return is monotone along the family.
The AAC formulation is also revised: whereas v2 paired an advanced-policy
actor update with TD regression of the policy-shaping Q-function,
Direct AAC in this version differentiates the advanced policy's return with respect to both
independent base coordinates, separating the policy-shaping Q-function
from its evaluator. These Direct updates are distinguished from
consistency-tracking (Chasing) updates and their combination.
The theoretical analysis, discrete- and continuous-action implementations,
and experiments are substantially expanded.
\end{quote}

%% file: sections/introduction.tex
\section{Introduction}

Policy gradient improves an explicit policy through a return gradient, while
Q-learning improves an implicit policy through its action-value function. 
Conventional actor--critic methods contain both objects but give them 
different roles: the actor is optimized and the critic evaluates it.

We study this distinction by using entropy in two ways. Augmenting the
reward with policy entropy defines the soft objective and its optimal policy
\citep{Ziebart2010,Neu2017,Geist2019,Haarnoja2018SAC}. Penalizing relative
entropy with respect to the current policy, by contrast, controls a single improvement step,
as in relative-entropy search and mirror-descent policy optimization
\citep{Peters2010REPS,Shani2020,Tomar2020MDPO,Abdolmaleki2018MPO}. We denote
the reward-level coefficient by $\alpha$ and the step-level coefficient by
$\beta$. Varying $\beta$ at fixed $\alpha$ traces a path from the current
policy to its soft-greedy policy, which we reparameterize by
$\lambda=\alpha/(\alpha+\beta)$ with endpoints $\lambda=0$ and $1$. We call
this the \emph{advanced-policy family}.

We apply policy gradient directly to the advanced policy, relaxing the
requirement that the action-value function evaluate the base policy and
treating the policy and action-value function as independent coordinates.
The Q coordinate directly shapes the policy rather than merely evaluating
the base actor.
Differentiating its objective yields updates for both coordinates, 
recovering soft policy gradient and an action-centered Q-learning-like update 
at the endpoints. This defines 
Advanced Actor--Critic (AAC), in which the connection between policy gradient
and Q-learning emerges from optimizing one composite policy, rather than from 
an imposed equivalence between them. At intermediate
$\lambda$, neither base coordinate need be meaningful on its own; 
the optimized object is their composite advanced policy. For continuous
actions, we fit a parametric actor to this generally implicit policy.

Our contributions are:
\begin{enumerate}[leftmargin=*,nosep]
  \item We introduce AAC, which differentiates the advanced policy objective 
  through independent policy and action-value coordinates, recovering soft
  policy gradient and Q-learning-like endpoint updates.
  \item We separate reward entropy from step-level KL regularization and give 
  a first-principles characterization of the advanced family's improvement, 
  fixed-point, and proximal properties.
  \item We provide discrete- and continuous-action implementations and
  evaluate endpoint and intermediate policies on control tasks.
\end{enumerate}

%% file: sections/related_work.tex
\section{Related Work}

Maximum-entropy RL augments return with policy entropy, yielding soft values,
Boltzmann improvement, and soft Q-learning
\citep{Ziebart2010,Geist2019,Haarnoja2018SAC}. Our action-inclusive values use
the same objective \citep{Wu2022UnderstandingPG}; this is a choice of notation that restores centered value--advantage identities, 
not a new soft-value definition.

KL-regularized policy iteration, relative-entropy policy search, and policy
mirror descent control the geometry of an individual improvement step
\citep{Peters2010REPS,Neu2017,Shani2020,Tomar2020MDPO,Abdolmaleki2018MPO}.
The consistent advanced-policy family is an entropic mirror-descent step with
reward entropy coefficient $\alpha$ kept distinct from the step-level KL coefficient $\beta$.
AAC moves beyond that consistent setting by treating policy and value as
independent coordinates. Chasing AAC is related to approximate regularized
policy iteration and MPO-style target projection: it repeatedly constructs an
improved target and tracks it with parametric policy/value representations.
Unlike an exact policy-space mirror-descent or KL-projection update, however,
its base and fast pairs remain separate parameterizations and are coupled through
a tracking update. Natural policy gradient and trust-region methods
provide related local geometric views
\citep{Amari1998,Kakade2001,Schulman2015TRPO,Schulman2017PPO}.

PGQL combines policy-gradient and Q-learning updates, while PCL minimizes soft
path-consistency residuals \citep{Odonoghue2017PGQ,Nachum2017PCL}.
\citet{Schulman2017Equivalence} decompose a soft Q-regression semi-gradient
into policy-gradient and value-regression components. Their Q-induced policy
resembles our $\lambda=1$ endpoint; AAC instead starts from the objective 
of an advanced policy composed from independent policy and Q coordinates, 
whose endpoints recover policy gradient and action-relative Q-learning-like updates. 
See also operator interpolations \citep{Wagner2013,Ghosh2020Operator}.

%% file: sections/preliminaries.tex
\section{Preliminaries}

Consider a discounted finite MDP
$(\mathcal S,\mathcal A,P,r,\gamma)$ with $\gamma\in(0,1)$, bounded rewards,
and initial distribution $\rho_0$ \citep{Puterman1994,Sutton2018}. We use the
unnormalized discounted occupancy
$\rho_\pi(s)=\E_\pi[\sum_{t\geq0}\gamma^t\mathbf1\{s_t=s\}]$, so that
$\eta(\pi)=\sum_{s,a}\rho_\pi(s)\pi(a|s)r(s,a)$. Without loss of
generality, we restrict $\mathcal S$ to states reachable with positive
probability from the support of $\rho_0$. The finite-action analysis assumes
full policy support, $\pi(a|s)>0$ for every $(s,a)$; therefore every
retained state has positive discounted occupancy under every such policy.
All logarithms and KL divergences are then well-defined, and $\pi$ can be represented by
finite softmax logits; boundary policies are understood as limits.

For temperature $\alpha>0$, we use the conventional soft action-value that
excludes the entropy of the conditioned current action:
\begin{equation}
\begin{aligned}
 Q_\alpha^\pi(s,a)
 &=r(s,a)+\gamma\E_{s'\sim P(\cdot|s,a),\,a'\sim\pi}
   [Q_\alpha^\pi(s',a')-\alpha\log\pi(a'|s')],\\
 V_\alpha^\pi(s)&=\E_{a\sim\pi}[Q_\alpha^\pi(s,a)-\alpha\log\pi(a|s)],\\
 A_\alpha^\pi(s,a)&=Q_\alpha^\pi(s,a)-V_\alpha^\pi(s).
\end{aligned}
\end{equation}
Unless stated otherwise, value, action-value, and advantage functions are understood 
to be the corresponding soft quantities. 
We omit $\alpha$ and $\pi$ when their meaning is clear.

%% file: sections/entropy_augmented_rl.tex
\section{Entropy-Augmented Policy Improvement}

We start from an entropy-augmented reward within the standard 
reinforcement-learning formulation:
\begin{equation}
\widetilde r_\pi(s,a)=r(s,a)-\alpha\log\pi(a|s).
\label{eq:augmented_reward}
\end{equation}
Its expectation is identical to adding $\alpha H(\pi(\cdot|s))$ at each
state. Let $\widetilde\eta(\pi)$ denote the discounted return under
$\widetilde r_\pi$. We now adopt an \emph{action-inclusive} convention by
assigning the entropy of the conditioned action to its action value
\citep{Wu2022UnderstandingPG}:
\begin{align}
\widetilde Q^\pi(s,a)&=Q_\alpha^\pi(s,a)-\alpha\log\pi(a|s),\\
\widetilde V^\pi(s)&=V_\alpha^\pi(s)
=\E_{a\sim\pi}[\widetilde Q^\pi(s,a)],\\
\widetilde A^\pi(s,a)&=\widetilde Q^\pi(s,a)-\widetilde V^\pi(s)
=A_\alpha^\pi(s,a)-\alpha\log\pi(a|s).
\label{eq:augmented_advantage}
\end{align}
These quantities are discounted sums of the entropy-augmented reward.
This is only a change in how the entropy term is included: it defines the same maximum-entropy
objective as the conventional $Q_\alpha^\pi$ above. We use the term canonical 
only in the limited sense that it restores the ordinary identities
$\widetilde V^\pi=\E_\pi[\widetilde Q^\pi]$,
$\widetilde A^\pi=\widetilde Q^\pi-\widetilde V^\pi$, and
$\E_{a\sim\pi}\widetilde A^\pi(s,a)=0$. We do not claim this convention as a
new value-function definition.

We use the following known entropy-regularized counterpart of the 
classical performance-difference lemma, stated in our action-inclusive 
notation \citep{KakadeLangford2002,Mei2020Softmax}. 
Because \(\widetilde r_\pi\) depends on the policy, comparing returns 
under different policies introduces an additional 
KL correction term.

\begin{aaclemma}[Entropy-regularized performance difference]
For two full-support policies $\pi$ and $\pi'$, their entropy-augmented returns
satisfy
\begin{equation}
\widetilde\eta(\pi')-\widetilde\eta(\pi)
=\sum_s\rho_{\pi'}(s)\sum_a\pi'(a|s)
\left[\widetilde A^\pi(s,a)
-\alpha\log\frac{\pi'(a|s)}{\pi(a|s)}\right].
\label{eq:soft_performance_difference}
\end{equation}
\end{aaclemma}

The policy-dependent-reward step underlying this correction is derived
explicitly in Appendix~\ref{app:soft_performance_difference}.

The same action-inclusive advantage is the exact soft policy-gradient
signal \citep{Neu2017,Wu2022UnderstandingPG}. For a differentiable
full-support policy $\pi_\theta$,
\begin{equation}
\nabla_\theta\widetilde\eta(\pi_\theta)
=\sum_s\rho_{\pi_\theta}(s)
\sum_a\pi_\theta(a|s)
\nabla_\theta\log\pi_\theta(a|s)
\widetilde A^{\pi_\theta}(s,a).
\label{eq:soft_policy_gradient_canonical}
\end{equation}
Thus $\widetilde A^\pi$ serves both as the centered statewise improvement
coordinate and as the policy-gradient critic. In unrestricted tabular policy
coordinates, it measures not simply how good an action is, but whether
its current probability should change: a positive value calls for increasing that
probability, and a negative value for decreasing it. A short variational
derivation from the preceding lemma is given in
Appendix~\ref{app:soft_policy_gradient}. In tabular
coordinates, $\widetilde A^\pi$ is a probability-consistency residual, which vanishes at
soft optimality.

For a fixed state, the bracketed expectation is maximized by
\begin{equation}
\pi_*(a|s)
\propto\pi(a|s)
\exp\!\left(\frac{\widetilde A^\pi(s,a)}{\alpha}\right)
\propto\exp\!\left(\frac{Q_\alpha^\pi(s,a)}{\alpha}\right).
\label{eq:soft_greedy}
\end{equation}
Here $\pi_*$ denotes the one-step soft-greedy policy computed from $Q_\alpha^\pi$;
it should not be confused with the globally optimal policy $\pi^*$. 
Although the first form appears as an exponential tilt of the base policy, 
the second shows that the dependence on the base policy cancels completely 
at the full soft-greedy endpoint; the cancellation and the resulting improvement
are explained in Appendix~\ref{app:advanced_policy_derivations}.

At a soft-optimal fixed point, $\pi=\pi_*=\pi^*$, equivalently
\begin{equation}
\widetilde A^{\pi^*}(s,a)=0
\quad\text{for all }(s,a).
\label{eq:soft_consistency}
\end{equation}
Under the finite, full-support, exact-value assumptions used here, the
converse also holds: vanishing $\widetilde A^\pi(s,a)$ for every state and
action makes $\pi$ a fixed point of soft policy improvement and hence
soft-optimal. This equation is the one-step form of the soft policy--value
consistency condition used by \citet{Nachum2017PCL}. Path Consistency Learning
treats its sampled multi-step residual as an error to minimize. Here the same
quantity is used as a centered policy-improvement signal and, below, as a
coordinate of the advanced-policy path. The distinction is one of use and
interpretation rather than a new optimality condition.

%% file: sections/advanced_policy.tex
\section{The Advanced-Policy Family}
\label{sec:advanced_policy}

Entropy augmentation prevents the soft-greedy policy from collapsing to a
deterministic one. It does not, however, preserve direct dependence on the
current policy: the $-\alpha\log\pi(a|s)$ term inside
$\widetilde A^\pi$ exactly cancels the multiplicative base policy. The
unconstrained statewise optimizer in Eq.~(\ref{eq:soft_greedy}) therefore
takes a full soft-greedy step, effectively forgetting where the update
started. Such a full step can move farther from the current policy than
its performance gain justifies, producing an unnecessarily concentrated 
policy target; imperfect value estimation can make this behavior still more
problematic. This motivates regularizing the improvement step itself: we seek
a partial optimizer that does not move too far from the current policy, yet
advances toward the same soft-greedy endpoint.

To make this partial improvement explicit, fix a state $s$ and write
$\pi_s=\pi(\cdot|s)$. We denote the original statewise improvement
objective by
\begin{equation}
\mathcal F_s(q)
=
\E_{a\sim q}[\widetilde A^\pi(s,a)]
-\alpha\KL(q\Vert\pi_s).
\label{eq:statewise_improvement_objective}
\end{equation}
Note that the KL term in $\mathcal F_s$ is not a proximal regularizer;
it arises from the policy dependence of the entropy-augmented reward when
comparing the two policies. We therefore introduce a separate penalty, with
coefficient $\beta$, to limit the size of the improvement step. Writing
$\widetilde A_\alpha^\pi=\widetilde A^\pi$ here to make the dependence on
the reward temperature explicit, we obtain
\begin{align}
&\widetilde\eta(\pi')-\widetilde\eta(\pi)
-\beta\sum_s\rho_{\pi'}(s)
\KL(\pi'(\cdot|s)\Vert\pi_s) \notag\\
&\quad=
\sum_s\rho_{\pi'}(s)\E_{a\sim\pi'(\cdot|s)}
\left[
\widetilde A_\alpha^\pi(s,a)
-(\alpha+\beta)\log\frac{\pi'(a|s)}{\pi(a|s)}
\right].
\label{eq:regularized_performance_difference}
\end{align}
Although the two KL contributions combine algebraically into the coefficient
$\alpha+\beta$, they play different roles. The $\alpha$ term is part of the
underlying soft objective: it determines the Bellman equations and the
soft-greedy endpoint, and remains even when no proximal penalty is imposed.
The $\beta$ term acts only on the improvement step, controlling how far the
new policy moves from the current one without changing either the underlying
return or its optimum. Maximizing the statewise integrand gives
\begin{equation}
\pi'_\beta(a|s)
\propto
\pi(a|s)
\exp\!\left(
\frac{\widetilde A_\alpha^\pi(s,a)}{\alpha+\beta}
\right)
\propto
\pi(a|s)^{\frac{\beta}{\alpha+\beta}}
\left[
\exp\!\left(\frac{Q_\alpha^\pi(s,a)}{\alpha}\right)
\right]^{\frac{\alpha}{\alpha+\beta}}.
\label{eq:kl_regularized_improvement}
\end{equation}

The second form makes the partial cancellation explicit: a power of the
base policy remains. The additional KL penalty thus controls 
how far the base policy is advanced toward its soft-greedy policy. 
This is an entropic mirror-descent update
\citep{Neu2017,Geist2019,Shani2020,Tomar2020MDPO}. For fixed $\alpha$, varying
$\beta$ changes the size of the step toward a fixed soft-greedy endpoint,
rather than changing the target soft objective itself.

The penalized form also has an equivalent trust-region interpretation:
it is the Lagrangian form of maximizing the original statewise objective
$\mathcal F_s$ under an appropriately chosen KL constraint. Specifically,
$\pi'_\beta$ maximizes $\mathcal F_s(q)$ subject to
\begin{equation}
\KL(q\Vert\pi_s)\leq
\KL(\pi'_\beta(\cdot|s)\Vert\pi_s).
\label{eq:induced_kl_constraint}
\end{equation}
Thus $\beta$ selects a point on a nested family of induced KL constraints. A
single shared $\beta$ generally induces a different radius at each state, so
Eq.~(\ref{eq:induced_kl_constraint}) should not be read as imposing one common
statewise radius. The closed-form derivation and the penalty--constraint
equivalence are given in Appendix~\ref{app:advanced_policy_derivations}.

Set $\lambda=\alpha/(\alpha+\beta)\in[0,1]$, with $\lambda=0$ understood as the
limit $\beta\to\infty$. This normalized coordinate makes the interpolation 
ratio explicit. The optimizer becomes
\begin{equation}
\pi'_\lambda(a|s)
\propto\pi(a|s)\exp\!\left(\frac{\lambda}{\alpha}\widetilde A^\pi(s,a)\right)
\propto\pi(a|s)^{1-\lambda}
\exp\!\left(\frac{\lambda}{\alpha}Q_\alpha^\pi(s,a)\right).
\label{eq:advanced_policy}
\end{equation}
We call this the \emph{advanced-policy family}, since each member partially 
advances the base policy toward the soft-greedy policy using the base policy's  
action-value evaluation. It traces a continuous path
that connects $\pi'_0=\pi$ to the one-step soft-greedy policy
$\pi'_1=\pi_*$. If $\pi=\operatorname{softmax}(u)$, the same construction is
especially simple:
\begin{equation}
\pi'_\lambda
=\operatorname{softmax}_a\!\left((1-\lambda)u(s,a)
+\frac{\lambda}{\alpha}Q_\alpha^\pi(s,a)\right).
\label{eq:logit_interpolation}
\end{equation}
It is therefore a straight interpolation in logit space, starting from the
current policy and ending at its soft-greedy improvement.

Under the consistency condition $Q=Q_\alpha^\pi$, this is a standard 
entropic mirror-descent policy-improvement path. Section~\ref{sec:algorithm} 
departs from this setting by relaxing that consistency condition and optimizing 
the advanced policy through independent policy and action-value coordinates. 
We first record the improvement, fixed-point, and proximal properties of 
the consistent family.

\subsection{Improvement and fixed points}

For fixed $s$, let
\begin{equation}
D_s(\lambda)=\KL(\pi'_\lambda(\cdot|s)\Vert\pi_s),
\qquad
F_s(\lambda)=\mathcal F_s(\pi'_\lambda(\cdot|s)).
\end{equation}
Standard log-partition differentiation shows that the induced radius and the
value of the original statewise objective are both nondecreasing along the path.
\begin{aacproposition}[Nested statewise improvement path]
For every $\lambda\in[0,1]$,
\begin{align}
\frac{d}{d\lambda}D_s(\lambda)
&=\frac{\lambda}{\alpha^2}
\Var_{a\sim\pi'_\lambda}[\widetilde A^\pi(s,a)]\geq0,
\label{eq:kl_distance_monotonicity}\\
\frac{d}{d\lambda}F_s(\lambda)
&=\frac{1-\lambda}{\alpha}
\Var_{a\sim\pi'_\lambda}[\widetilde A^\pi(s,a)]\geq0.
\label{eq:statewise_monotonicity}
\end{align}
Except when $\widetilde A^\pi(s,\cdot)$ is constant, the first inequality is
strict for $\lambda>0$ and the second for $\lambda<1$.
\end{aacproposition}
These identities explain the constraint interpretation: increasing
$\lambda$ (equivalently, decreasing $\beta$) relaxes the induced KL constraint
and cannot reduce the original statewise objective along this constrained-optimum 
path. Since $F_s(0)=0$, the
performance-difference identity gives the central guarantee. The
log-partition calculation is in Appendix~\ref{app:advanced_policy_derivations}.

\begin{aaccorollary}[Improvement over the base policy]
For every $\lambda\in(0,1]$,
\begin{equation}
\widetilde\eta(\pi'_\lambda)\geq\widetilde\eta(\pi).
\label{eq:base_improvement}
\end{equation}
By the corresponding Bellman policy-improvement argument, the same statewise
inequality yields
$Q_\alpha^{\pi'_\lambda}(s,a)\geq Q_\alpha^\pi(s,a)$ and
$V_\alpha^{\pi'_\lambda}(s)\geq V_\alpha^\pi(s)$.
\end{aaccorollary}

This is the standard soft policy-improvement guarantee applied along the
partial KL-regularized path
\citep{Haarnoja2018SAC}; the Bellman argument is given
in Appendix~\ref{app:advanced_policy_derivations}.

Equation~(\ref{eq:base_improvement}) compares every member with the common base
policy. It does \emph{not} order different family members: changes in
$\rho_{\pi'_\lambda}$ can make
$\widetilde\eta(\pi'_\lambda)$ nonmonotone in $\lambda$.
Appendix~\ref{app:advanced_path_analysis} decomposes the return derivative
into statewise improvement and occupancy redistribution, and examines the
resulting path empirically. No algorithm or claim below relies on
monotonicity along the family.

Define $T_\lambda\pi=\pi'_\lambda$ using the exact $Q_\alpha^\pi$.
The fixed-point condition follows directly from the exponential form. We state it
explicitly because the correct optimality statement concerns fixed points,
rather than simultaneous optimality of all policies generated from an
arbitrary base; this agrees with the standard optimality theory of regularized
MDPs \citep{Neu2017,Geist2019}.
\begin{aacproposition}[Common optimal fixed point]
For every fixed $\lambda\in(0,1]$,
\begin{equation}
T_\lambda\pi=\pi
\quad\Longleftrightarrow\quad
\widetilde A^\pi(s,a)=0\ \text{for all }(s,a),
\label{eq:common_fixed_point}
\end{equation}
and hence all nontrivial operators share the same soft-optimal fixed point.
The statement excludes $\lambda=0$, for which $T_0$ is the identity.
\end{aacproposition}
Appendix~\ref{app:common_fixed_point} gives the fixed-point argument and
clarifies why it does not imply simultaneous optimality of an arbitrary base
and its advanced policy. An advanced policy produced from a nonoptimal base
can nevertheless be optimal; for example, the $\lambda=1$ endpoint may
already be optimal after a single soft-greedy step. Optimality of
$\pi'_\lambda$ therefore does not imply optimality of its base policy.

\subsection{Proximal limit}

It is convenient locally to write
$\epsilon=1/(\alpha+\beta)$ and
$\pi'_\epsilon\propto\pi e^{\epsilon\widetilde A^\pi}$. At the current policy,
\begin{align}
\left.\frac{d\pi'_\epsilon(a|s)}{d\epsilon}\right|_{0}
&=\pi(a|s)\widetilde A^\pi(s,a),
\label{eq:proximal_tangent}\\
\left.\frac{d\widetilde\eta(\pi'_\epsilon)}{d\epsilon}\right|_{0}
&=\sum_{s,a}\rho_\pi(s)\pi(a|s)
\bigl(\widetilde A^\pi(s,a)\bigr)^2\geq0.
\label{eq:proximal_ascent}
\end{align}
Appendix~\ref{app:proximal_limit} proves
Eqs.~\eqref{eq:proximal_tangent} and~\eqref{eq:proximal_ascent}.
Thus the advanced-policy path initially follows the natural policy-gradient 
direction and ends at the soft-greedy policy. More precisely, for a differentiable policy
class, its Fisher-metric projection into parameter space has velocity
\begin{equation}
\left.\frac{d\theta}{d\epsilon}\right|_0
=F(\theta)^\dagger\nabla_\theta\widetilde\eta(\pi_\theta).
\label{eq:fisher_weighted_projection}
\end{equation}
Here $F(\theta)$ is the Fisher information matrix and $F(\theta)^\dagger$
denotes its Moore--Penrose pseudoinverse. This is an exact representation of
Eq.~\eqref{eq:proximal_tangent} when the tangent lies in the policy class;
otherwise it is only its local projection.
Appendix~\ref{app:fisher_projection} derives
Eq.~\eqref{eq:fisher_weighted_projection} in parameter coordinates.

Under the tabular full-support assumptions, Eq.~\eqref{eq:proximal_ascent}
also gives a result about local maxima in policy space. If $\pi$ is not soft-optimal, then
$\widetilde A^\pi$ is nonzero at some reachable state--action pair, and the
right-hand side of that equation is strictly positive.
Hence an arbitrarily small advanced-policy step improves
$\widetilde\eta$: the entropy-augmented objective has no suboptimal interior
local maxima.  This statement concerns direct optimization over policies; it
does not exclude stationary points induced by a restricted neural
parameterization.

%% file: sections/algorithm.tex
\section{Advanced Actor--Critic}
\label{sec:algorithm}

AAC provides two independently selectable updates for its advanced policy.
\emph{Direct AAC} differentiates the return of that policy, giving the
policy-gradient-to-Q-learning-like path. \emph{Chasing AAC} instead tracks
the advanced policy and its evaluator to promote consistency. Enabling both
updates yields \emph{Combined AAC}. The three modes share the same
advanced-policy construction but differ in how the base coordinates are
updated.

All modes are organized around the composed \emph{advanced policy} defined by
the independent coordinates $\pi_\theta$ and $Q_\phi$,
\begin{equation}
\pi'_{\theta,\phi}(a|s)
=\frac{1}{Z_{\theta,\phi}(s)}
\pi_\theta(a|s)^{1-\lambda}
\exp\!\left(\frac{\lambda}{\alpha}Q_\phi(s,a)\right).
\label{eq:aac_policy}
\end{equation}
Under $Q_\phi=Q_\alpha^{\pi_\theta}$, this recovers the consistent family of
Section~\ref{sec:advanced_policy}; during training no mode imposes that
equality. Different coordinate pairs can represent the same $\pi'$, so
improving it need not improve either coordinate separately.

For implementation, we introduce a fast pair
$(\pi_f,Q_f)$. The fast actor provides a tractable approximation to $\pi'$, 
while the fast critic evaluates whichever policy is actually executed, 
either $\pi'$ itself or its approximation $\pi_f$. Direct AAC improves the
advanced policy by differentiating its objective through the base coordinates.
Chasing AAC instead moves the base coordinates toward the advanced policy and
its action-value function; as the base pair changes, the advanced policy is
redefined and can continue to change through repeated updates.

\subsection{Direct AAC: advanced-policy optimization}

Direct AAC differentiates $\widetilde\eta(\pi'_{\theta,\phi})$ through both
base coordinates. In the exact formulation, the required policy-gradient
signal is the canonical soft advantage of the advanced policy,
\begin{equation}
\widetilde A^{\pi'}(s,a)
=
Q_\alpha^{\pi'}(s,a)
-\alpha\log\pi'(a|s)
-V_\alpha^{\pi'}(s).
\label{eq:aac_advanced_advantage}
\end{equation}
Differentiating the normalizer $Z_{\theta,\phi}(s)$ in
Eq.~\eqref{eq:aac_policy} introduces a centering term in each coordinate
gradient, but these terms vanish because
$\E_{a\sim\pi'}[\widetilde A^{\pi'}(s,a)]=0$. Thus
\begin{align}
\nabla_\theta\widetilde\eta(\pi')
&=(1-\lambda)\E_{\rho_{\pi'},\pi'}
[\nabla_\theta\log\pi_\theta(a|s)\,
\widetilde A^{\pi'}(s,a)],
\label{eq:actor_channel}\\
\nabla_\phi\widetilde\eta(\pi')
&=\frac{\lambda}{\alpha}\E_{\rho_{\pi'},\pi'}
[\nabla_\phi Q_\phi(s,a)\,
\widetilde A^{\pi'}(s,a)].
\label{eq:q_channel}
\end{align}
With exact evaluation and on-policy state weighting, these are the 
gradients of $\widetilde\eta(\pi')$. The derivation is in
Appendix~\ref{app:aac_gradient_derivations}.

\paragraph{Endpoint interpretation.}
At $\lambda=0$, $Q_\phi$ drops out of $\pi'$, and the Q-channel gradient
vanishes. Equation~\eqref{eq:actor_channel} then reduces to the soft policy
gradient because $\pi'=\pi_\theta$ and
$\widetilde A^{\pi'}=\widetilde A^{\pi_\theta}$.
At $\lambda=1$, $\pi_\theta$ drops out of $\pi'$, and the actor-channel
gradient vanishes. Since
$\alpha\log\pi'=Q_\phi-\alpha\log Z$, the canonical soft advantage becomes
\begin{equation}
\widetilde A^{\pi'}(s,a)
=\bigl(Q_\alpha^{\pi'}(s,a)-Q_\phi(s,a)\bigr)
-\E_{a'\sim\pi'}
[Q_\alpha^{\pi'}(s,a')-Q_\phi(s,a')].
\label{eq:centered_q_residual}
\end{equation}
Replacing $Q_\alpha^{\pi'}$ by a Bellman target gives a centered TD residual.
Unlike soft Q-learning, this gradient identifies only action-relative values:
$Q_\phi+c(s)$ represents the same policy. The stationary tabular condition
therefore matches soft Q-learning up to this gauge
(Appendix~\ref{app:action_relative_q}). This corresponds to the
policy-changing component of the soft-Q gradient decomposition of
\citet{Schulman2017Equivalence}; their full regression also fits a state-value
component.

\paragraph{Implementation.}
For discrete actions, Eq.~\eqref{eq:aac_policy} and its action expectations are
tractable. A categorical $\pi_f$ tracks $\pi'$ by minimizing
$\KL(\pi'\Vert\pi_f)$, equivalently categorical cross-entropy with the exact
probabilities of $\pi'$. The discrete Q channel uses an exact action sum and
a coverage mixture around $\pi'$; implementation details are given in
Appendix~\ref{app:aac_implementation_details}.

For continuous actions, Eq.~\eqref{eq:aac_policy} generally lacks tractable 
normalization and direct sampling. Sampling-based methods or richer density 
approximations provide alternative implementations (Appendix~\ref{app:projection_chasing_interpretation}). For computational simplicity, 
however, we use squashed-Gaussian base and fast policies and fit $\pi_f$ by reverse KL:
\begin{equation}
\KL(\pi_f\Vert\pi')
=\lambda\,\E_{a\sim\pi_f}\left[
\log\pi_f(a|s)-\frac{Q_\phi(s,a)}{\alpha}\right]
+(1-\lambda)\,\KL(\pi_f\Vert\pi_\theta)
+\log Z_{\theta,\phi}(s).
\label{eq:reverse_kl}
\end{equation}
For fixed base coordinates and temperature, this fits $\pi_f$ to increase
its expected $Q_\phi$ value and entropy while penalizing reverse KL
to the base policy.
The normalizer is constant in the fast parameters; fitting uses only
$\pi_f$ samples (Appendix~\ref{app:projection_chasing_interpretation}).
Continuous Direct updates use centered action samples from $\pi_f$ at replay
states, giving practical approximations to the exact gradients
(Appendix~\ref{app:aac_implementation_details}).

\subsection{Chasing AAC: advanced-pair consistency}

If $Q_\phi=Q_\alpha^{\pi_\theta}$, each nontrivial $\pi'$ improves on
$\pi_\theta$. This motivates moving the base pair toward the advanced policy
and its value,
\begin{equation*}
\pi_\theta\ \longrightarrow\ \pi'_{\theta,\phi},
\qquad
Q_\phi\ \longrightarrow\ Q_\alpha^{\pi'_{\theta,\phi}}.
\end{equation*}
Because $\pi'_{\theta,\phi}$ is redefined after every base update, repeated
chasing can itself move the advanced policy forward when $\lambda>0$, giving
an approximate form of regularized policy iteration with interleaved updates.
At $\lambda=0$, however, $\pi'_0=\pi_\theta$, so this actor-side improvement
mechanism vanishes.

In our experiments, we use matching parameterizations, $\pi_f=\pi_{\theta_f}$
 and $Q_f=Q_{\phi_f}$, enabling Polyak interpolation for chasing.
\begin{equation}
\theta\leftarrow(1-\tau)\theta+\tau\theta_f,\qquad
\phi\leftarrow(1-\tau)\phi+\tau\phi_f.
\label{eq:chasing}
\end{equation}
Here $\tau$ controls the tracking timescale. For continuous actions, 
as $Q_\phi$ tracks the soft evaluator of
$\pi_f$, Eq.~\eqref{eq:reverse_kl} has a SAC-like interpretation
with reverse-KL regularization toward the moving base policy.
Since the fast pair is itself a
moving target, chasing does not in general guarantee either a reduction in
policy--value inconsistency or an improvement in the return of $\pi'$.
Under exact tabular evaluation and consistency, however, its nontrivial fixed
point satisfies the soft-optimality condition.

\subsection{Direct, chasing, and combined AAC}

Setting $\tau=0$ disables chasing and yields \emph{Direct AAC}; it is the
mode whose endpoint gradients give soft policy gradient and an
action-relative Q-learning-like update. Setting the learning rates for the 
Direct-AAC gradients to zero yields
\emph{Chasing AAC}. Both retain fast-policy fitting and
evaluation. Enabling both gives \emph{Combined AAC}. Chasing encourages 
consistency between the coordinates, whereas Direct AAC allows them to remain 
independent while optimizing the composed policy.
Their update directions therefore need not align. Strong chasing can also
reduce the effective expressivity of the implicit policy class under function
approximation, as the modality diagnostic in
Appendix~\ref{app:chasing_expressivity} illustrates. We make no general
performance-ordering claim.

With Adam, Direct AAC scales its two learning rates by $1-\lambda$ and
$\lambda$; the Q-channel rate absorbs $1/\alpha$. In our Combined AAC
experiments, the direct and chasing rates are separate hyperparameters;
their relative strengths depend on the environment-specific settings. Replay collection uses a separate
exploration mixture, and time-limit truncations are bootstrapped.

\begin{algorithm}[t]
\caption{Advanced Actor--Critic (AAC)}
\label{alg:aac}
\begin{algorithmic}[1]
\STATE Initialize base $(\pi_\theta,Q_\phi)$, fast $(\pi_f,Q_f)$,
targets, and replay buffer $\mathcal B$.
\STATE Select Direct AAC, Chasing AAC, or both.
\STATE \textbf{Repeat}
\STATE \hspace{\algorithmicindent}Construct $\pi'$ and collect a transition
using $\pi'$ if tractable, otherwise $\pi_f$; store it in $\mathcal B$.
\STATE \hspace{\algorithmicindent}Sample a mini-batch from $\mathcal B$.
\STATE \hspace{\algorithmicindent}Fit $\pi_f$ to $\pi'$; update $Q_f$ by soft TD.
\STATE \hspace{\algorithmicindent}\textbf{Direct:} if enabled, update $(\theta,\phi)$ using the practical
forms of Eqs.~\eqref{eq:actor_channel}--\eqref{eq:q_channel}
(Appendix~\ref{app:aac_implementation_details}).
\STATE \hspace{\algorithmicindent}\textbf{Chasing:} move $(\theta,\phi)$ toward
$(\theta_f,\phi_f)$ by Eq.~\eqref{eq:chasing}, if enabled.
\end{algorithmic}
\end{algorithm}

%% file: sections/experiments.tex
\section{Experiments}
\label{sec:experiments}

We test whether AAC learns at endpoint and intermediate values of $\lambda$,
how its three update modes compare, and how chasing affects
implicit-policy modality. We report executable-policy returns: the
categorical advanced policy for discrete actions and the fast actor for
continuous actions.

\paragraph{Protocol.}
The control tasks are Acrobot, CartPole, MountainCar, LunarLander,
Pendulum, HalfCheetah, and Walker2d. We compare Direct, Chasing, and Combined
AAC with environment-specific budgets (Table~\ref{tab:main_control_summary}).
The five-seed results use ten evaluation episodes per seed and
sample standard deviations across seed means. The Walker2d comparison
reports 80 one-million-step runs: five seeds at
$\lambda\in\{0,0.1,0.3,0.5,0.7,1\}$ for Direct, and
$\lambda\in\{0.1,0.3,0.5,0.7,1\}$ for Chasing and Combined. Settings were fixed after limited
preliminary screening; method-specific learning rates differ, so these
comparisons do not isolate chasing alone. Full experimental settings, parameter sweeps, and per-seed results
are given in Appendices~\ref{app:preliminary_control}
and~\ref{app:walker2d_results}.

\begin{table*}[t]
\centering
\small
\caption{Executable-policy returns at $\lambda=.5$ (mean $\pm$ sample
standard deviation across training seeds; ten evaluation episodes per seed).
All cells use five seeds except LunarLander Combined ($n=3$).
Budgets and method settings vary by environment. Full experimental settings 
and run-level results are provided in 
Appendices~\ref{app:bookkeeping_control} and~\ref{app:walker2d_results}.}
\label{tab:main_control_summary}
\resizebox{\textwidth}{!}{%
\begin{tabular}{llrrrrr}
\toprule
Action & Environment & Steps & $\alpha$ & Direct AAC & Chasing AAC & Combined AAC\\
\midrule
Discrete & Acrobot-v1 & 50k & .001 & $-77.6\pm5.8$ & $-101.7\pm20.3$ & $-88.3\pm7.3$\\
Discrete & MountainCar-v0 & 500k & .001 & $-119.7\pm44.9$ & $-128.9\pm43.9$ & $-101.0\pm4.8$\\
Discrete & CartPole-v1 & 50k & .001 & $373.0\pm179.2$ & $412.8\pm77.5$ & $425.1\pm122.8$\\
Discrete & LunarLander-v3 & 500k & .001 & $236.5\pm39.9$ & $270.4\pm29.4$ & $168.0\pm147.7$ (3)\\
Continuous & Pendulum-v1 & 50k & .2 & $-178.2\pm11.5$ & $-177.5\pm10.8$ & $-177.8\pm11.8$\\
Continuous & HalfCheetah-v5 & 200k & .2 & $5748\pm847$ & $6492\pm951$ & $6380\pm1202$\\
Continuous & Walker2d-v5 & 1M & .2 & $3422\pm1436$ & $2778\pm560$ & $3290\pm924$\\
\bottomrule
\end{tabular}
}
\end{table*}

\paragraph{Control results.}
Table~\ref{tab:main_control_summary} summarizes the available results at
$\lambda=0.5$, with the smaller LunarLander Combined sample explicitly marked.
The discrete sweeps show learning at intermediate values and no uniform
ordering of update modes across $\lambda$. At the representative value,
Direct has the highest mean on Acrobot, whereas Combined has the highest
mean on MountainCar. With continuous actions, the Gaussian fast-policy
projection supports learning on Pendulum and both locomotion tasks.
Pendulum mainly serves as an implementation check; HalfCheetah supplies a
five-seed comparison at 200k steps, with Chasing having the highest mean
in the current results.

\begin{figure*}[t]
\centering
\includegraphics[width=\textwidth]{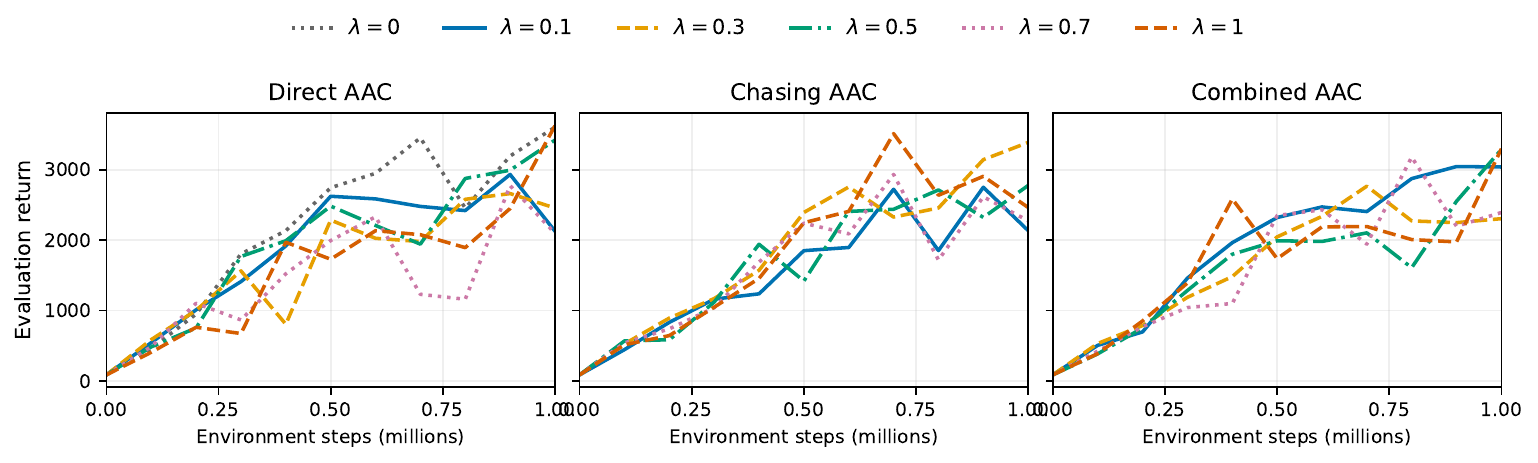}
\caption{Walker2d-v5 learning curves at $\lambda\in\{0,0.1,0.3,0.5,0.7,1\}$.
Panels separate the three update modes; colors and line styles identify
$\lambda$. Lines show mean executable-policy returns over five training
seeds, with ten deterministic episodes per evaluation. Only Direct is reported at $\lambda=0$. Curves with across-seed standard
deviations appear in Appendix~\ref{app:walker2d_results}.}
\label{fig:walker2d_main_learning_curves}
\end{figure*}

Figure~\ref{fig:walker2d_main_learning_curves} shows the longer Walker2d
comparison. Direct learns at both endpoints, reaching final means of 3605
and 3627 at $\lambda=0$ and $1$, and also learns at intermediate values.
The highest means are attained by Chasing at $\lambda=0.3$, Combined at
$\lambda=0.1,0.7$, and Direct at $\lambda=0.5,1$. Some Chasing and Combined settings show
large variation across seeds, including near-zero final scores; all seeds are
retained. At $\lambda=1$, all five Direct seeds exceed a final return of 2600.
Combined varies more across seeds: four seeds score between 3881
and 4268, while one scores 34. This spread, not shown in the mean-only
curves, highlights the importance of the individual-seed results
reported in Appendix~\ref{app:walker2d_results}.
These results support learning across the family but do not establish
statistical superiority among modes. We do not include matched SAC, TD3,
or PPO runs. Appendix~\ref{app:published_baselines} provides published
baseline results for context; differences in environments and evaluation
protocols prevent a controlled comparison with AAC.

\paragraph{Implicit-policy modality.}
Although the executable policy is a single Gaussian, the implicit
advanced policy can represent multiple action modes through its
independent Q coordinate. We test whether chasing reduces this
multimodality by encouraging policy--value consistency.
A 10-state bimodal diagnostic uses exact policy evaluation on a
discretized action grid to measure this effect before projection
onto the executable policy. Strong chasing reduces the fraction of multimodal state policies
from 96.7\% to 3.0\% (Figure~\ref{fig:advanced_policy_modality}), while the
mean executable-return difference remains small under the shared
single-Gaussian projection. This illustrates an expressivity--consistency
trade-off without establishing a general performance ordering.
Appendix~\ref{app:chasing_expressivity} gives the setup and paired results.

\begin{figure}[t]
\centering
\small
\begin{tabular}{@{}lrl@{}}
Direct AAC, $\tau=0$ & 96.7\% & \rule{0.435\linewidth}{1.35ex}\\[0.4ex]
Chasing, $\tau=10^{-3}$ & 44.7\% & \rule{0.201\linewidth}{1.35ex}\\[0.4ex]
Chasing, $\tau=10^{-2}$ & 3.0\% & \rule{0.014\linewidth}{1.35ex}
\end{tabular}
\caption{Fraction of states whose implicit advanced policy is multimodal in
the exact 10-state bimodal diagnostic ($30$ paired initializations,
$\lambda=.7$). Stronger chasing reduces the modality of $\pi'$ before the
common single-Gaussian fast-policy projection.}
\label{fig:advanced_policy_modality}
\end{figure}

%% file: sections/conclusion.tex
\section{Conclusion}

Separating reward entropy from step-level KL regularization yields an
advanced-policy path from a base policy to its soft-greedy update. Every
nontrivial member improves over the base under exact evaluation, though the
members need not be ordered by return. AAC relaxes policy--value consistency
and differentiates the composed policy's return through independent policy and
Q coordinates, recovering soft policy gradient and an action-centered
Q-learning-like update at the endpoints. A tractable fast policy extends the
method to continuous actions. Control experiments show both successes and
failures, while an exact diagnostic shows that consistency can restrict
implicit-policy modality. Matched baselines and additional seeds remain
necessary for broader empirical assessment.

%% file: sections/ai_assistance.tex
\section*{AI Assistance Disclosure}

Large language models assisted with literature discovery, research
discussion, drafting and revising proofs, checking mathematical
arguments, providing feedback on experimental design, implementing
and debugging code, interpreting experimental results, preparing
figures and tables, and drafting and editing the manuscript.
The author directed the research and reviewed the AI-assisted text,
code, analyses, and mathematical claims, and takes responsibility
for the final content of the paper.

%% file: sections/reproducibility.tex
\section*{Reproducibility Statement}

The appendix provides proofs and additional derivations for the theoretical
results. Implementation details are given in
Appendix~\ref{app:aac_implementation_details}, and experimental settings
and per-seed results in Appendices~\ref{app:preliminary_control}
and~\ref{app:walker2d_results}. The anonymous supplementary package includes 
the AAC implementation, run configurations, evaluation records underlying 
the reported experiments, and scripts for recomputing the reported summary 
statistics and Walker2d learning curves. Its README documents installation, 
execution, and which data are included
and what their limitations are.

%% file: sections/appendix.tex
\section{Proof Sketches and Additional Details}

This appendix provides proofs, implementation details, and additional
experiments. For the main theoretical results, start with
Appendices~\ref{app:soft_performance_difference}--\ref{app:fisher_projection}:
these derive the soft policy identities, the advanced-policy path, its
common fixed point, and its initial relation to natural policy gradient.
Appendix~\ref{app:surrogate_policy_gradients} gives a further local connection
between fitting a fixed advanced target and soft policy gradient.

For the algorithm, Appendices~\ref{app:aac_gradient_derivations}
and~\ref{app:action_relative_q} explain the exact AAC gradients, the practical
approximations, and the Q channel's freedom to shift values without changing
the policy. Appendix~\ref{app:claim_scope} summarizes what the improvement
and optimality results do and do not establish.
Appendices~\ref{app:advanced_path_analysis} and~\ref{app:chasing_expressivity}
then examine two separate questions: how return changes along the policy
path, and how chasing affects the implicit policy's action distribution.
Detailed control results and settings are in
Appendices~\ref{app:preliminary_control} and~\ref{app:walker2d_results};
Appendix~\ref{app:published_baselines} provides published baseline results
for context, rather than a comparison under matched conditions.

\subsection{Soft performance difference}
\label{app:soft_performance_difference}

The policy dependence of the augmented reward requires one step beyond a
literal substitution into the classical performance-difference lemma.  We use
the decomposition
\begin{equation}
 -\alpha\log\pi'(a| s)
 =-\alpha\log\pi(a| s)
  -\alpha\log\frac{\pi'(a| s)}{\pi(a| s)},
 \label{eq:entropy_reward_decomposition}
\end{equation}
which separates a reward shared by the two return terms from the correction
caused by changing the policy.  Expanding the two entropy-augmented returns
therefore gives
\begin{align}
 \widetilde\eta(\pi')-\widetilde\eta(\pi)
 &=\sum_{s,a}\rho_{\pi'}(s)\pi'(a| s)
       \widetilde r_\pi(s,a)
   -\sum_{s,a}\rho_\pi(s)\pi(a| s)
       \widetilde r_\pi(s,a) \notag\\
 &\quad
   -\alpha\sum_{s,a}\rho_{\pi'}(s)\pi'(a| s)
       \log\frac{\pi'(a| s)}{\pi(a| s)}.
 \label{eq:soft_pdl_decomposition}
\end{align}
The first two sums now use the same fixed reward
$\widetilde r_\pi$.  The ordinary performance-difference lemma can thus be
applied to them, yielding
\begin{align}
 \widetilde\eta(\pi')-\widetilde\eta(\pi)
 &=\sum_{s,a}\rho_{\pi'}(s)\pi'(a| s)
 \left[
   \widetilde A^\pi(s,a)
   -\alpha\log\frac{\pi'(a| s)}{\pi(a| s)}
 \right],
\end{align}
which proves Eq.~\eqref{eq:soft_performance_difference}.  Notice that the
classical lemma is not applied to two different policy-dependent rewards; it
is applied only after Eq.~\eqref{eq:entropy_reward_decomposition} has made the
reward in the first two sums identical.  An equivalent soft
performance-difference identity appears in \citet[Lemma~25]{Mei2020Softmax},
under normalized state occupancy; exchanging the two policies and
reversing the sign of the return difference gives the form above.
Its integrand may
equivalently be written as
$A_\alpha^\pi(s,a)-\alpha\log\pi'(a| s)$.

\subsection{Soft policy gradient in canonical form}
\label{app:soft_policy_gradient}

We derive Eq.~\eqref{eq:soft_policy_gradient_canonical} directly from the
soft performance-difference identity.  Fix $\pi=\pi_\theta$ and perturb the
second policy to $\pi'=\pi_{\theta+\delta\theta}$.  At
$\delta\theta=0$, the first variation of
Eq.~\eqref{eq:soft_performance_difference} is
\begin{align}
 \delta\widetilde\eta
 &=\sum_s\delta\rho_\pi(s)
      \underbrace{\sum_a\pi(a| s)\widetilde A^\pi(s,a)}_{=0}
   +\sum_s\rho_\pi(s)\sum_a\delta\pi(a| s)
       \widetilde A^\pi(s,a)\notag\\
 &\quad
   -\alpha\sum_s\rho_\pi(s)\sum_a\delta\pi(a| s).
 \label{eq:soft_pg_first_variation}
\end{align}
The occupancy variation vanishes because the canonical advantage is centered,
and the last term vanishes by policy normalization.  Using
$\delta\pi=\pi\,\delta\log\pi$ leaves
\begin{equation}
 \delta\widetilde\eta
 =\sum_{s,a}\rho_\pi(s)\pi(a| s)
   \delta\log\pi(a| s)\widetilde A^\pi(s,a),
\end{equation}
and identifying the coefficient of the arbitrary perturbation $\delta\theta$ gives
Eq.~\eqref{eq:soft_policy_gradient_canonical}.  This argument also explains
why no explicit derivative of the discounted occupancy remains in the policy
gradient: its coefficient is the zero conditional mean of
$\widetilde A^\pi$.

\subsection{Statewise optimization, constraints, and monotonicity}
\label{app:advanced_policy_derivations}

We first find the best action distribution at a fixed state, then explain
how the KL penalty controls its distance from the base policy. Finally,
we show why each policy on the path improves over that base.
Throughout, fix a state $s$ and write
$\widetilde A(a)=\widetilde A^\pi(s,a)$ and $p(a)=\pi(a|s)$.
The distribution $q$ is the policy we optimize at this state;
$p$ and $\widetilde A$ remain fixed.

\paragraph{Finding the maximizing policy.}
Define the statewise improvement objective
\begin{equation}
 \mathcal F_s(q)
 =\sum_aq(a)\left[\widetilde A(a)-\alpha\log\frac{q(a)}{p(a)}\right].
 \label{eq:app_statewise_improvement_objective}
\end{equation}
The statewise integrand in Eq.~\eqref{eq:regularized_performance_difference} is
$\mathcal F_s(q)-\beta\KL(q\Vert p)$, which we maximize over probability
distributions $q$. To enforce $\sum_a q(a)=1$, introduce a multiplier
$\nu$ and form the Lagrangian
\begin{equation}
 \mathcal L(q,\nu)
 =\sum_aq(a)\widetilde A(a)
  -(\alpha+\beta)\sum_aq(a)\log\frac{q(a)}{p(a)}
  +\nu\left(\sum_aq(a)-1\right).
\end{equation}
Because $p$ has full support and $\alpha+\beta>0$, the objective is strictly
concave, so its maximizer is unique. At that maximizer, every action has
positive probability: if $q(a)=0$, moving a small mass $\delta$ to $a$
from an action with positive probability increases the objective by
$-(\alpha+\beta)\delta\log\delta+O(\delta)>0$ for sufficiently small
$\delta>0$. We can therefore set the derivative with respect to each
$q(a)$ to zero, giving
\begin{equation}
 \widetilde A(a)-(\alpha+\beta)
 \left(1+\log\frac{q(a)}{p(a)}\right)+\nu=0.
\end{equation}
Solving for $q(a)$ gives a quantity proportional to
$p(a)e^{\widetilde A(a)/(\alpha+\beta)}$. Requiring the probabilities
to sum to one yields
\begin{equation}
 q(a)=\frac{p(a)e^{\widetilde A(a)/(\alpha+\beta)}}
 {\sum_b p(b)e^{\widetilde A(b)/(\alpha+\beta)}},
\end{equation}
which proves Eq.~\eqref{eq:kl_regularized_improvement}.

\paragraph{The soft-greedy endpoint.}
When $\beta=0$, only the reward-entropy coefficient $\alpha$ remains.
We can check the optimizer directly and calculate its improvement. Let
\begin{equation}
 Z_A(s)=\sum_a p(a)e^{\widetilde A(a)/\alpha},
 \qquad q_*(a)=\frac{p(a)e^{\widetilde A(a)/\alpha}}{Z_A(s)}.
\end{equation}
For any action distribution $q$, the original
statewise objective can be rewritten as a constant minus a KL divergence:
\begin{equation}
 \mathcal F_s(q)
 =\alpha\log Z_A(s)-\alpha\KL(q\Vert q_*).
 \label{eq:statewise_kl_completion}
\end{equation}
The term $\alpha\log Z_A(s)$ does not depend on $q$, and the KL divergence
is nonnegative, with equality only at $q=q_*$. Thus $q_*$ uniquely
maximizes the objective, with value $\alpha\log Z_A(s)$.
To see that this value is nonnegative, use
$\E_{a\sim p}[\widetilde A(a)]=0$ and Jensen's inequality:
\begin{equation}
 Z_A(s)=\E_{a\sim p}[e^{\widetilde A(a)/\alpha}]
 \geq e^{\E_p[\widetilde A]/\alpha}=1.
\end{equation}
Because $p(a)>0$ for every action, equality requires all
$\widetilde A(a)$ to be equal. Their mean is zero, so they must all be
zero. Otherwise, the maximum is strictly greater than the base value
$\mathcal F_s(p)=0$.
The optimizer is also the familiar softmax of $Q_\alpha^\pi$:
using $\widetilde A^\pi=Q_\alpha^\pi-V_\alpha^\pi-\alpha\log p$ gives
\begin{align}
 p(a)e^{\widetilde A^\pi(s,a)/\alpha}
 &=e^{-V_\alpha^\pi(s)/\alpha}
   e^{Q_\alpha^\pi(s,a)/\alpha},\label{eq:base_policy_cancellation}\\
 q_*(a)&=
 \frac{e^{Q_\alpha^\pi(s,a)/\alpha}}
      {\sum_b e^{Q_\alpha^\pi(s,b)/\alpha}}.
\end{align}
The $-\alpha\log p(a)$ term cancels the factor $p(a)$, while the
state-dependent value $V_\alpha^\pi(s)$ cancels on normalization.
Writing $Z_Q(s)=\sum_a e^{Q_\alpha^\pi(s,a)/\alpha}$, the two
normalizing constants satisfy
\begin{equation}
 Z_A(s)=e^{-V_\alpha^\pi(s)/\alpha}Z_Q(s),
 \qquad
 \alpha\log Z_A(s)=\alpha\log Z_Q(s)-V_\alpha^\pi(s).
\end{equation}
Thus $\alpha\log Z_A(s)$ is the increase from $V_\alpha^\pi(s)$ to
its one-step soft-greedy Bellman update.

Choosing $q_*$ at every state and applying the soft performance-difference
identity gives
\begin{equation}
 \widetilde\eta(q_*)-\widetilde\eta(\pi)
 =\alpha\sum_s\rho_{q_*}(s)\log Z_A(s)\geq0.
 \label{eq:full_soft_greedy_improvement}
\end{equation}
The return increases strictly if $Z_A(s)>1$ at a state visited with positive
discounted occupancy. Our state space includes only reachable states,
and full action support gives $\rho_{q_*}(s)>0$ at each one. Hence the
returns are equal only when $\widetilde A^\pi$ vanishes everywhere and
$q_*=\pi$. If unreachable states are included, this conclusion applies
only to reachable states.

\paragraph{From a KL penalty to a KL constraint.}
Let $q_\beta$ be the policy found above: it maximizes
$\mathcal F_s(q)-\beta D(q)$, where $D(q)=\KL(q\Vert p)$ and $\beta\geq0$.
Set $\delta_\beta=D(q_\beta)$, the distance this policy actually moves
from $p$. We show that $q_\beta$ also maximizes the original objective
among policies no farther from $p$:
\begin{equation}
 \max_q\ \mathcal F_s(q)
 \quad\text{subject to}\quad D(q)\leq\delta_\beta.
 \label{eq:induced_trust_region_problem}
\end{equation}
For any $q$ satisfying $D(q)\leq\delta_\beta$, the defining optimality
of $q_\beta$ gives
$\mathcal F_s(q)-\beta D(q)\leq
\mathcal F_s(q_\beta)-\beta D(q_\beta)$. Rearranging yields
\begin{align}
 \mathcal F_s(q)
 &\leq \mathcal F_s(q_\beta)
       +\beta\bigl(D(q)-D(q_\beta)\bigr)\\
 &\leq \mathcal F_s(q_\beta).
\end{align}
Thus no policy within this KL bound has a larger objective than
$q_\beta$. The optimizer attains the bound by construction.
This argument uses only $D(q)\leq D(q_\beta)$ and $\beta\geq0$;
it does not require monotonicity of the KL distance along the policy path.
That separate property, proved in Eq.~\eqref{eq:app_kl_distance_derivative}
below, shows that the induced radius is nondecreasing in $\lambda=\alpha/(\alpha+\beta)$
(and hence nonincreasing in $\beta$).
When $\beta=0$, $q_0=q_*$ already maximizes the objective without a
constraint, so any radius at least $D(q_0)$ preserves this optimizer.

\paragraph{How the KL distance and statewise objective change along the path.}
For $\lambda=\alpha/(\alpha+\beta)$, the normalizing constant determines
both derivatives. Write its logarithm as
\begin{equation}
 \psi_s(\lambda)
 =\log\sum_a p(a)e^{\lambda \widetilde A(a)/\alpha}.
\end{equation}
Differentiating the finite sum gives
\begin{equation}
 \psi_s'(\lambda)=\frac{1}{\alpha}\E_{\pi'_\lambda}[\widetilde A],
 \qquad
 \psi_s''(\lambda)=\frac{1}{\alpha^2}
 \Var_{\pi'_\lambda}[\widetilde A].
 \label{eq:log_partition_derivatives}
\end{equation}
Since
$\log(\pi'_\lambda/p)=\lambda \widetilde A/\alpha-\psi_s(\lambda)$,
\begin{equation}
 D_s(\lambda)=\lambda\psi_s'(\lambda)-\psi_s(\lambda),
\end{equation}
and therefore
\begin{equation}
 D_s'(\lambda)=\lambda\psi_s''(\lambda)
 =\frac{\lambda}{\alpha^2}\Var_{\pi'_\lambda}[\widetilde A]\geq0.
 \label{eq:app_kl_distance_derivative}
\end{equation}
This proves Eq.~\eqref{eq:kl_distance_monotonicity}. Likewise,
\begin{align}
 F_s(\lambda)
 &=\E_{\pi'_\lambda}[\widetilde A]-\alpha D_s(\lambda)\\
 &=\alpha\bigl((1-\lambda)\psi_s'(\lambda)
               +\psi_s(\lambda)\bigr),
\end{align}
so
\begin{equation}
 F_s'(\lambda)=\alpha(1-\lambda)\psi_s''(\lambda)
 =\frac{1-\lambda}{\alpha}\Var_{\pi'_\lambda}[\widetilde A].
\end{equation}
This proves Eq.~\eqref{eq:statewise_monotonicity}. Both derivatives are
nonnegative on $[0,1]$. For nonconstant $\widetilde A$, the variance is
positive, so $D_s'(\lambda)>0$ for $\lambda>0$ and
$F_s'(\lambda)>0$ for $\lambda<1$.

\paragraph{From statewise improvement to return and value improvement.}
At the base policy, $F_s(0)=0$. Since $F_s$ is nondecreasing,
$F_s(\lambda)\geq0$ at every state. By
Eq.~\eqref{eq:soft_performance_difference},
\begin{equation}
 \widetilde\eta(\pi'_\lambda)-\widetilde\eta(\pi)
 =\sum_s\rho_{\pi'_\lambda}(s)F_s(\lambda)\geq0,
\end{equation}
which proves Eq.~\eqref{eq:base_improvement}. To show that the soft value
functions also improve, first average the base Q-function and the new
policy's entropy at each state:
\begin{equation}
 W_\lambda(s)
 =\sum_a\pi'_\lambda(a| s)
   \left[Q_\alpha^\pi(s,a)
         -\alpha\log\pi'_\lambda(a| s)\right].
\end{equation}
Using the definitions of $\widetilde A^\pi$ and $F_s$ gives
\begin{equation}
 W_\lambda(s)=V_\alpha^\pi(s)+F_s(\lambda)
 \geq V_\alpha^\pi(s).
 \label{eq:w_lambda_improvement}
\end{equation}
Hold $q=\pi'_\lambda$ fixed during policy evaluation. Define the soft
Bellman operators on action values and state values, respectively, by
\begin{align}
 (\mathcal T_q^Q Q)(s,a)
 &=r(s,a)+\gamma\E_{s'|s,a}\E_{a'\sim q(\cdot|s')}
 [Q(s',a')-\alpha\log q(a'|s')],\\
 (\mathcal T_q^V V)(s)
 &=\E_{a\sim q(\cdot|s)}
 [r(s,a)-\alpha\log q(a|s)+\gamma\E_{s'|s,a}V(s')].
 \label{eq:app_soft_evaluation_operators}
\end{align}
For either operator, subtracting its values at two inputs cancels the
reward and entropy terms, leaving $\gamma$ times an expectation of their
difference. Both operators therefore preserve pointwise inequalities and
are $\gamma$-contractions in the supremum norm.
Consequently,
\begin{align}
 Q_\alpha^\pi(s,a)
 &=r(s,a)+\gamma\E_{s'| s,a}[V_\alpha^\pi(s')]\notag\\
 &\leq r(s,a)+\gamma\E_{s'| s,a}[W_\lambda(s')]
 =(\mathcal T_q^Q Q_\alpha^\pi)(s,a).
\end{align}
Repeated application thus gives a nondecreasing sequence converging to
$Q_\alpha^q$. Starting from $Q_\alpha^\pi$ yields
\begin{equation}
 Q_\alpha^\pi
 \leq\lim_{k\to\infty}(\mathcal T_q^Q)^k Q_\alpha^\pi
 =Q_\alpha^{\pi'_\lambda}.
\end{equation}
Likewise, the soft state-value evaluation operator obeys
\begin{equation}
 (\mathcal T_q^V V_\alpha^\pi)(s)
 =W_\lambda(s)\geq V_\alpha^\pi(s),
\end{equation}
and its repeated application gives
$V_\alpha^{\pi'_\lambda}\geq V_\alpha^\pi$.  This proof uses only the
statewise inequality $F_s(\lambda)\geq0$; it does not require actual return to
be monotone in $\lambda$.

\subsection{Common fixed point}
\label{app:common_fixed_point}

For $\lambda>0$, $T_\lambda\pi=\pi$ implies that
$e^{\lambda\widetilde A^\pi(s,a)/\alpha}$ is action-independent at every
state. Thus $\widetilde A^\pi(s,a)$ is action-independent. Its expectation
under $\pi$ is zero, so $\widetilde A^\pi=0$, the soft Bellman optimality
condition. The converse follows immediately. This argument concerns fixed
points of the operators and does not imply that optimality of a single
$T_\lambda\pi$ forces optimality of the base policy.

\subsection{Proximal tangent and ascent}
\label{app:proximal_limit}

We parameterize the advanced-policy path by $\epsilon=\lambda/\alpha$,
with $\alpha$ held fixed.

Fix a state and abbreviate $\widetilde A(a)=\widetilde A^\pi(s,a)$. For
$\pi'_\epsilon(a)\propto\pi(a)e^{\epsilon \widetilde A(a)}$, the normalizer satisfies
\begin{align*}
 Z_s(\epsilon)&=\sum_b\pi(b)e^{\epsilon\widetilde A(b)},\\
 \frac{d}{d\epsilon}\log Z_s(\epsilon)
 &=\frac{\sum_b\pi(b)e^{\epsilon\widetilde A(b)}\widetilde A(b)}
         {Z_s(\epsilon)}
 =\E_{b\sim\pi'_\epsilon}[\widetilde A(b)].
\end{align*}
Differentiating the normalized policy therefore gives
\begin{align}
 \frac{d\pi'_\epsilon(a)}{d\epsilon}
 &=\pi'_\epsilon(a)
 \left(\widetilde A(a)-\E_{b\sim\pi'_\epsilon}[\widetilde A(b)]\right).
\end{align}
At $\epsilon=0$, the centering identity
$\E_{a\sim\pi}[\widetilde A^\pi(s,a)]=0$ therefore yields
Eq.~\eqref{eq:proximal_tangent}. Equivalently, the Taylor argument makes the
vanishing normalization term explicit:
\begin{align}
 Z_s(\epsilon)
 &=\sum_a\pi(a| s)e^{\epsilon \widetilde A(a)}
   =1+\epsilon\E_\pi[\widetilde A]+O(\epsilon^2)
   =1+O(\epsilon^2),\\
 \pi'_\epsilon(a| s)
 &=\pi(a| s)+\epsilon\pi(a| s)\widetilde A(a)+O(\epsilon^2).
 \label{eq:proximal_taylor_expansion}
\end{align}
Here the base policy and its advantage are held fixed while $\epsilon$ varies;
re-evaluating $\widetilde A$ along the path would define a different derivative.
Applying the entropy-regularized policy-gradient
identity to this distributional direction gives
\begin{align}
 \left.\frac{d\widetilde\eta(\pi'_\epsilon)}{d\epsilon}
 \right|_{0}
 &=\sum_s\rho_\pi(s)\sum_a
   \left.\frac{d\pi'_\epsilon(a| s)}{d\epsilon}\right|_0
   \widetilde A^\pi(s,a)\\
 &=\sum_{s,a}\rho_\pi(s)\pi(a| s)
   \bigl(\widetilde A^\pi(s,a)\bigr)^2,
\end{align}
which proves Eq.~\eqref{eq:proximal_ascent}. These are standard
exponential-family and policy-gradient identities; their role here is to
identify the local geometry of the finite advanced-policy path.

\subsection{Fisher-metric projection and natural gradient}
\label{app:fisher_projection}

We project the policy-space tangent derived in
Appendix~\ref{app:proximal_limit} into parameter space to obtain
Eq.~\eqref{eq:fisher_weighted_projection}.
For a restricted differentiable policy class, let
$g_\theta(s,a)=\nabla_\theta\log\pi_\theta(a| s)$ and seek the parameter
velocity that best represents the distributional tangent
$d\log\pi/d\epsilon=\widetilde A$ in weighted least squares:
\begin{equation}
 \min_v\ \frac12\sum_{s,a}\rho_\pi(s)\pi(a| s)
       \bigl(g_\theta(s,a)^\top v-\widetilde A(s,a)\bigr)^2.
 \label{eq:tangent_projection_ls}
\end{equation}
Its normal equation is
\begin{equation}
 F(\theta)v
 =\sum_{s,a}\rho_\pi(s)\pi(a| s)
   g_\theta(s,a)\widetilde A(s,a)
 =\nabla_\theta\widetilde\eta(\pi_\theta),
\end{equation}
where
\begin{equation}
 F(\theta)=\sum_{s,a}\rho_\pi(s)\pi(a| s)
 g_\theta(s,a)g_\theta(s,a)^\top
\end{equation}
is the discounted Fisher information matrix. Choosing the minimum Euclidean
norm solution of the normal equation gives
\begin{equation}
 \left.\frac{d\theta}{d\epsilon}\right|_0
 =F(\theta)^\dagger\nabla_\theta\widetilde\eta(\pi_\theta),
 \label{eq:advanced_tangent_natural_gradient}
\end{equation}
with the Moore--Penrose inverse covering redundant parameterizations.
Other solutions differ by a vector in the null space of $F$, which does not
change the policy tangent on state--action pairs with positive weight. The
identity is exact when the tangent is representable; otherwise it is its
Fisher-weighted projection onto the policy model's tangent space.  This is the
precise sense in which the advanced-policy path initially follows natural
policy gradient \citep{Kakade2001}.

\subsection{Policy gradients of advanced-policy surrogates}
\label{app:surrogate_policy_gradients}

This section asks how fitting a policy to a fixed advanced target relates
to improving its return. At the starting policy, reverse-KL fitting gives
the same gradient direction as soft policy gradient. This explains a local
connection between target fitting and return optimization; it does not
derive the parameter-averaging update used by Chasing AAC. Let
$p=\pi_{\theta_0}$, $\widetilde A=\widetilde A^p$, and
\begin{equation}
 q_\epsilon(a| s)
 =\frac{p(a| s)e^{\epsilon \widetilde A(s,a)}}{Z_s(\epsilon)}.
\end{equation}
Freeze both this target and the state weights $\rho_p$, and consider the
reverse-KL surrogate
\begin{equation}
 \mathcal J_{\mathrm{KL}}(\theta)
 =-\sum_s\rho_p(s)
   \KL\!\left(\pi_\theta(\cdot| s)\Vert
                    q_\epsilon(\cdot| s)\right).
\end{equation}
Since
$\log q_\epsilon=\log p+\epsilon \widetilde A-\log Z_s(\epsilon)$,
policy normalization removes the state-only terms and gives, exactly for any
finite $\epsilon$,
\begin{align}
 \left.\nabla_\theta\mathcal J_{\mathrm{KL}}(\theta)
 \right|_{\theta=\theta_0}
 &=\epsilon\sum_{s,a}\rho_p(s)p(a| s)
   \nabla_\theta\log\pi_\theta(a| s)|_{\theta_0}
   \widetilde A(s,a)\\
 &=\epsilon\nabla_\theta\widetilde\eta(\pi_\theta)
   |_{\theta_0}.
 \label{eq:reverse_kl_surrogate_pg}
\end{align}
Thus a reverse-KL projection toward any point on this fixed exponential path
has the same initial direction as soft policy gradient, differing only by the
scale $\epsilon$.  This is a local statement: after $\pi_\theta$ moves away
from $p$, the KL objective also contains a pull toward the fixed target and is
not identical to the performance objective.  If the state weights are instead
allowed to depend on $\pi_\theta$, differentiating them adds a term weighted
by $\KL(p\Vert q_\epsilon)=O(\epsilon^2)$ at the starting point.

The choice of fitting loss matters. For example, consider the negative
squared-error objective
\begin{equation}
 \mathcal J_2(\theta)
 =-\frac12\sum_{s,a}\rho_p(s)
   \bigl(\pi_\theta(a| s)-q_\epsilon(a| s)\bigr)^2
\end{equation}
and the expansion
$q_\epsilon=p+\epsilon p\widetilde A+O(\epsilon^2)$ yield
\begin{align}
 \left.\nabla_\theta\mathcal J_2(\theta)\right|_{\theta_0}
 &=\epsilon\sum_{s,a}\rho_p(s)p(a| s)\widetilde A(s,a)
   \nabla_\theta\pi_\theta(a| s)|_{\theta_0}
   +O(\epsilon^2)\\
 &=\epsilon\sum_{s,a}\rho_p(s)p(a| s)^2\widetilde A(s,a)
   \nabla_\theta\log\pi_\theta(a| s)|_{\theta_0}
   +O(\epsilon^2).
\end{align}
The extra action-probability factor changes the update direction even though
the target policy is the same.

\subsection{AAC gradient and projection derivations}
\label{app:aac_gradient_derivations}

Taking logarithms of Eq.~\eqref{eq:aac_policy} gives
\begin{equation}
 \log\pi'=(1-\lambda)\log\pi_\theta
 +\frac{\lambda}{\alpha}Q_\phi-\log Z_{\theta,\phi}.
\end{equation}
The log-partition derivatives are expectations under $\pi'$, so
\begin{align}
 \nabla_\theta\log\pi'(a| s)
 &=(1-\lambda)\left(
   \nabla_\theta\log\pi_\theta(a| s)
   -\E_{b\sim\pi'}[\nabla_\theta\log\pi_\theta(b| s)]\right),\\
 \nabla_\phi\log\pi'(a| s)
 &=\frac{\lambda}{\alpha}\left(
   \nabla_\phi Q_\phi(s,a)
   -\E_{b\sim\pi'}[\nabla_\phi Q_\phi(s,b)]\right).
\end{align}
Insert these expressions into the soft policy-gradient identity
\begin{equation}
 \nabla\widetilde\eta(\pi')
 =\E_{\rho_{\pi'},\pi'}
 [\nabla\log\pi'(a| s)\widetilde A^{\pi'}(s,a)].
\end{equation}
Because $\E_{a\sim\pi'}\widetilde A^{\pi'}(s,a)=0$, expanding the centered
score terms yields Eqs.~\eqref{eq:actor_channel}--\eqref{eq:q_channel}.
With an approximate critic or a modified discrete sampling measure, the
estimated signal is instead centered under the measure used by the
implementation.

For continuous actions, the projection loss and its implementation are
derived in Appendix~\ref{app:projection_chasing_interpretation}.

\subsubsection{Practical surrogates and fast-pair evaluation}
\label{app:aac_implementation_details}

For discrete actions, let the fast-critic entropy-augmented action value be
\begin{equation}
\widetilde Q_{f,\pi'}(s,a)
= Q_f(s,a)-\alpha\log\pi'(a| s).
\label{eq:implemented_canonical_q}
\end{equation}
The subscript distinguishes it from the exact
$\widetilde Q^{\pi'}=Q_\alpha^{\pi'}-\alpha\log\pi'$: the fast critic
generally evaluates $\pi_f$ rather than the implicit $\pi'$.
When $\pi_f=\pi'$ and $Q_f=Q_\alpha^{\pi'}$, centering under
$\mu=\pi'$ recovers $\widetilde A^{\pi'}$. To improve action coverage we use
\begin{equation}
\mu=(1-\varepsilon)\pi'+\varepsilon\Unif(\mathcal A),\qquad
\widetilde A^\mu_{f,\pi'}(s,a)
=\widetilde Q_{f,\pi'}(s,a)
-\E_{a'\sim\mu(\cdot| s)}
[\widetilde Q_{f,\pi'}(s,a')],
\label{eq:discrete_surrogate}
\end{equation}
with $\varepsilon=0.1$. The exact gradient has $\mu=\pi'$; the mixture is a
coverage surrogate at intermediate $\lambda$. The discrete Q-coordinate update
is
\begin{equation}
\Delta\phi
\propto
\sum_{a\in\mathcal A}\mu(a| s)\,
\widetilde A^\mu_{f,\pi'}(s,a)\,\nabla_\phi Q_\phi(s,a).
\label{eq:discrete_q_update}
\end{equation}
The centering removes a uniform action shift.
At $\lambda=1$, under exact evaluation and independent tabular
Q coordinates, Proposition~\ref{prop:full_support_q_gauge} shows
that any full-support weighting $\mu$, including the uniform
mixture above, preserves the stationary condition
$Q_\phi(s,a)=Q_\alpha^{\pi'}(s,a)+c(s)$.
The same weighting must be used for both centering and the
action-wise update; the learning dynamics may nevertheless differ.

For continuous actions, the executable actor is $\pi_f$ and the fast
critic is trained by soft TD evaluation of that policy:
\begin{equation}
y=r+\gamma(1-d)
\bigl(\bar Q_f(s',a')-\alpha\log\bar\pi_f(a'| s')\bigr),
\qquad a'\sim\bar\pi_f,
\label{eq:critic_target}
\end{equation}
Here $y$ is the one-step soft TD target used to train $Q_f(s,a)$,
with gradients stopped through the target. The transition provides reward
$r$ and next state $s'$, and $d$ indicates true termination (not a time-limit
truncation). The networks $\bar\pi_f$ and $\bar Q_f$ are slowly updated
target copies of $\pi_f$ and $Q_f$, respectively, obtained by Polyak
averaging their parameters; $a'$ is sampled from $\bar\pi_f(\cdot|s')$.
Thus $Q_f$ estimates $Q_\alpha^{\pi_f}$, rather than the
generally intractable $Q_\alpha^{\pi'}$. With bounded actions and bounded
$Q_\phi$, the normalizer and expectations used here are finite; on an
unbounded domain, $\exp(Q_\phi/\alpha)$ need not be integrable.

For the continuous Direct update, sample $K\geq2$ independent actions
$a_i\sim\pi_f(\cdot|s)$ at each replay state ($K=4$ by default).
Define the unnormalized implicit log density and centered signal by
\begin{align}
 \ell_{\theta,\phi}(s,a)
 &=(1-\lambda)\log\pi_\theta(a|s)+\frac{\lambda}{\alpha}Q_\phi(s,a),\\
 z_i&=Q_f(s,a_i)-\alpha\ell_{\theta,\phi}(s,a_i),\qquad
 \widehat{\widetilde A}_i=\frac{K}{K-1}\left(z_i-\frac1K\sum_{j=1}^Kz_j\right).
 \label{eq:continuous_centered_surrogate}
\end{align}
The unknown state-only normalizer cancels on centering. The factor
$K/(K-1)$ makes this equivalent to subtracting the mean of the other
$K-1$ samples, avoiding the shrinkage from including a sample in its own
baseline. The actor and Q surrogates average
$\widehat{\widetilde A}_i\nabla_\theta\log\pi_\theta(a_i|s)$ and
$\widehat{\widetilde A}_i\nabla_\phi Q_\phi(s,a_i)$, respectively, with actions and
signals detached; the channel scaling is handled as described in the main
text. These estimate centered updates under $\pi_f$ and replay state
weights. With the theoretical channel factors restored, their expectations
recover the exact Direct gradients when
$\pi_f=\pi'$, exact evaluation, and the appropriate on-policy state
weighting; otherwise they are practical surrogates.

\subsubsection{Continuous-policy approximation and chasing interpretation}
\label{app:projection_chasing_interpretation}
The continuous advanced policy in Eq.~\eqref{eq:aac_policy} is defined
independently of the fast-policy family. Its unnormalized density permits
MCMC sampling or importance-weighted expectation estimates with suitable
proposals; rejection sampling additionally requires a valid envelope.
Particle approximations or richer variational density models are other
possibilities. These approaches have their own computational and
approximation costs and do not automatically yield exact samples or gradients.
We choose a single squashed-Gaussian fast policy for computational simplicity;
this is an implementation choice, not a restriction on the AAC construction.

Substituting
Eq.~\eqref{eq:aac_policy} into the definition of reverse KL gives
\begin{equation*}
 \KL(\pi_f\Vert\pi')
 =\E_{a\sim\pi_f}\left[
 \log\pi_f(a|s)-(1-\lambda)\log\pi_\theta(a|s)
 -\frac{\lambda}{\alpha}Q_\phi(s,a)\right]
 +\log Z_{\theta,\phi}(s).
\end{equation*}
Splitting $\log\pi_f=(1-\lambda)\log\pi_f+\lambda\log\pi_f$
and regrouping yields Eq.~\eqref{eq:reverse_kl}.
The implementation minimizes $\alpha\KL(\pi_f\Vert\pi')$ after dropping
$\alpha\log Z_{\theta,\phi}(s)$; for fixed $\alpha>0$, this rescaling
preserves the minimizers. A restricted squashed-Gaussian fast-policy family
need not represent the implicit policy exactly, so the projection need not
reproduce its action expectations.
This also affects subsequent Direct updates: action regions assigned little
probability by $\pi_f$ are sampled infrequently, so learning signals may
concentrate on the modes favored by the projection. This does not imply
that other modes disappear from the implicit policy.
For fixed base coordinates and temperature, the normalizer term is constant
with respect to the fast-policy parameters. In Eq.~\eqref{eq:reverse_kl},
the $\lambda$-weighted term fits $\pi_f$ to increase
$\E_{a\sim\pi_f}[Q_\phi(s,a)/\alpha]+\mathcal H(\pi_f(\cdot|s))$
with $Q_\phi$ held fixed, while the $(1-\lambda)$-weighted term keeps
$\pi_f$ close to the base policy in reverse KL.
This identity holds whether or not policy--value consistency is satisfied.

A complementary local result appears in
Appendix~\ref{app:surrogate_policy_gradients}: when the fixed target is built
from exact base-policy evaluation and state weights are held at the base
occupancy, reverse-KL fitting starts in the soft policy-gradient direction
at the base policy. That result concerns target fitting, not the
parameter-averaging step used by chasing.

Under chasing, $Q_\phi$ tracks $Q_f$, which estimates
$Q_\alpha^{\pi_f}$ by the soft TD update above. To the extent that this
tracking and evaluation are accurate, the first term has the form of a
SAC actor objective scaled by $\lambda/\alpha$, with the second term
providing reverse-KL regularization toward the moving base policy.
This interpretation does not identify the full algorithm with SAC:
tracking is approximate and the KL penalty remains for $\lambda<1$.
Without chasing, $Q_\phi$ remains an independent coordinate and need not
evaluate $\pi_f$, although the same algebraic decomposition applies.

\subsection{Action-relative Q learning and the policy-invariant gauge}
\label{app:action_relative_q}

We make precise the relation between the $\lambda=1$ Q channel and soft
Q-learning. Fix a state $s$ and write
\begin{equation}
 D(a)=Q_\alpha^{\pi'}(s,a)-Q_\phi(s,a).
\end{equation}
For an action distribution $\mu(\cdot| s)$, define the centered update
signal
\begin{equation}
 h_\mu(a)=\mu(a| s)
 \left(D(a)-\E_{b\sim\mu}[D(b)]\right).
 \label{eq:mu_centered_q_signal}
\end{equation}

The centered signal has no component along the common-shift direction:
\begin{equation}
 \sum_a h_\mu(a)
 =\E_{a\sim\mu}[D(a)]
 -\E_{b\sim\mu}[D(b)]\sum_a\mu(a|s)=0.
 \label{eq:zero_sum_gauge_signal}
\end{equation}
Thus a tabular update $\Delta Q_\phi(s,a)=\eta h_\mu(a)$ with a common
scalar step size preserves the arithmetic mean of the action values at
$s$, introducing no drift along the additive gauge direction.

\begin{aacproposition}[Full-support stationary condition]
\label{prop:full_support_q_gauge}
Suppose the action set is finite, $\mu(a| s)>0$ for every action, and the
tabular values $Q_\phi(s,a)$ can be varied independently. Then
$h_\mu(a)=0$ for every action if and only if there exists a scalar $c(s)$ such
that
\begin{equation}
 Q_\phi(s,a)=Q_\alpha^{\pi'}(s,a)+c(s)
 \quad\text{for every }a.
 \label{eq:q_gauge_equivalence}
\end{equation}
Consequently, $Q_\phi(s,\cdot)$ and
$Q_\alpha^{\pi'}(s,\cdot)$ induce the same Boltzmann policy.
\end{aacproposition}

\begin{proof}
Because $\mu(a| s)>0$, Eq.~\eqref{eq:mu_centered_q_signal} vanishes for
every action exactly when
$D(a)=\E_{b\sim\mu}[D(b)]$ for every $a$. Thus $D$ is constant over actions,
which is equivalent to Eq.~\eqref{eq:q_gauge_equivalence}. The final statement follows because a state-dependent common
shift cancels from the Boltzmann normalizer:
\begin{equation}
 \frac{e^{(Q(a)+c)/\alpha}}{\sum_b e^{(Q(b)+c)/\alpha}}
 =\frac{e^{Q(a)/\alpha}}{\sum_b e^{Q(b)/\alpha}}.
\end{equation}
\end{proof}

The proposition concerns the representation needed for control. It does not
assert that the soft Bellman equation itself has a gauge freedom: for fixed
rewards and discount, that equation fixes the absolute value level. Standard
soft Q-learning learns this particular representative. The AAC endpoint
instead operates on the equivalence class
\begin{equation}
 Q(s,\cdot)\sim Q(s,\cdot)+c(s)\mathbf 1,
\end{equation}
because every member represents the same policy.
This gauge freedom reflects the policy-gradient origin of the
AAC Q channel: its objective depends on $Q_\phi$ through the
policy it defines, rather than through its absolute value level.
A state-dependent additive shift leaves that policy unchanged
and is therefore invisible to the exact objective.
If an absolute critic is desired, an additional gauge condition, for example
\begin{equation}
 \E_{a\sim\nu}[Q_\phi(s,a)]
 =\E_{a\sim\nu}[Q_\alpha^{\pi'}(s,a)]
\end{equation}
for any full-support $\nu$, sets $c(s)=0$ and recovers the evaluated Q-function.

The zero-sum property of the raw signal need not survive coordinate-wise
adaptive scaling or updates through shared nonlinear parameters, so the
common value level can drift in practical optimization. If needed, a weak
gauge-fixing regularizer can penalize the squared difference in the
mean-value condition above, using a detached estimate from $Q_f$ for the
reference value. This is an optional stabilization mechanism; with shared
parameters it can also affect action-relative values and does not guarantee
that drift is eliminated.
No explicit gauge-fixing regularizer was used in the
Direct AAC experiments reported here.

Proposition~\ref{prop:full_support_q_gauge} also shows that the
action-relative stationary condition is unchanged by the choice
of any full-support weighting distribution $\mu$. This separates
two issues: the stationary condition determines $Q_\phi$ only up
to a state-dependent additive constant, but does not uniquely
determine the weighting of action-wise updates. Different choices
of $\mu$ can nevertheless produce different update directions
and learning dynamics.

The exact AAC derivation fixes the update weighting by specifying
$\mu=\pi'$, while retaining the additive gauge freedom in $Q_\phi$.
This weighting follows from differentiating the performance of
the composed policy, rather than being chosen independently.
Since the soft policy has full support, it is sufficient for
the stationary condition in Proposition~\ref{prop:full_support_q_gauge}.
By comparison, sampled Q-learning weights state--action residuals implicitly
through its behavior and replay distribution, while an exact fitted update may
choose another full-support weighting such as the uniform distribution. These
choices have different conditioning and function-approximation projections,
even when their tabular policy fixed point agrees.

Our discrete implementation uses
\begin{equation}
 \mu_\varepsilon=(1-\varepsilon)\pi'
 +\varepsilon\Unif(\mathcal A).
\end{equation}
At $\lambda=1$, Proposition~\ref{prop:full_support_q_gauge} shows that this
modification preserves the action-relative stationary condition while giving
low-probability actions an update weight that is not too small. At intermediate
$\lambda$, however, replacing $\pi'$ by $\mu_\varepsilon$ changes the exact
policy gradient and must be understood as a practical coverage and
conditioning surrogate. With shared nonlinear parameters, even full support
does not prevent cancellations in parameter space, so the proposition should
not be read as a global convergence result for neural approximation.

\subsection{Scope of the improvement and optimality claims}
\label{app:claim_scope}

Under exact evaluation, Eq.~\eqref{eq:base_improvement} compares each
advanced policy with the same base policy; it does not say that return
increases with $\lambda$. For $\lambda\in(0,1]$, the operators instead
share a soft-optimal fixed point
(Eq.~\eqref{eq:common_fixed_point}, Appendix~\ref{app:common_fixed_point}),
which does not mean that an arbitrary base and its advanced policies
are simultaneously optimal. An advanced policy can be optimal even when
its base is not. Appendix~\ref{app:advanced_path_analysis} explains why
return can decrease along the path and measures this behavior in experiments.

\subsection{Return variation along the advanced-policy path}
\label{app:advanced_path_analysis}

Why can return decrease along the path when the statewise objective
increases? We separate the return derivative into statewise improvement
and changes in state occupancy, then examine the path empirically.

\subsubsection{Derivative decomposition}
\label{app:advanced_path_derivative}

Consider a finite discounted MDP with $\alpha>0$, $\gamma\in(0,1)$,
and fixed initial distribution $\rho_0$. Fix a full-support base policy
$\pi$ and its exact soft evaluation $Q_\alpha^\pi$ throughout the path.
Write $\widetilde\eta_\lambda=\widetilde\eta(\pi'_\lambda)$ and
$\rho_\lambda=\rho_{\pi'_\lambda}$. For $\lambda\in[0,1]$, the return derivative decomposes as
\begin{equation}
\begin{aligned}
 \frac{d\widetilde\eta_\lambda}{d\lambda}
 ={}&\underbrace{\alpha(1-\lambda)\sum_s\rho_\lambda(s)
 \Var_{a\sim\pi'_\lambda(\cdot|s)}
 \left[\log\frac{\pi_*(a|s)}{\pi(a|s)}\right]}_{L(\lambda):\ \text{statewise improvement}}\\
 &+\underbrace{\sum_{s,a}\rho_\lambda(s)\pi'_\lambda(a|s)u_\lambda(s,a)
 \left[(\bar G_\lambda-I)
 \bigl(A_\alpha^\pi-\alpha\log\pi'_\lambda\bigr)\right](s,a)
 }_{C(\lambda):\ \text{occupancy redistribution}}.
\end{aligned}
\label{eq:path_return_derivative_decomposition}
\end{equation}
The first term is nonnegative, whereas the second has no general sign.
The soft-greedy policy $\pi_*$, centered score $u_\lambda$, and
discounted Green's operator $\bar G_\lambda$ are defined below.

\paragraph{Policy path and its derivative.}
Define the base policy's soft-greedy policy and their log ratio by
\begin{equation}
 \pi_*(a|s)
 =\frac{\exp(Q_\alpha^\pi(s,a)/\alpha)}
 {\sum_b\exp(Q_\alpha^\pi(s,b)/\alpha)},\qquad
 \ell(s,a)=\log\frac{\pi_*(a|s)}{\pi(a|s)}.
 \label{eq:path_log_ratio}
\end{equation}
The soft-greedy policy need not be the globally optimal policy.
The advanced family is
\begin{equation}
 \pi'_\lambda(a|s)
 =\frac{1}{Z_\lambda(s)}\pi(a|s)\exp(\lambda\ell(s,a)),
 \qquad Z_\lambda(s)=\sum_a\pi(a|s)\exp(\lambda\ell(s,a)).
\end{equation}
Define the centered log-ratio score
\begin{equation}
 u_\lambda(s,a)
 =\ell(s,a)-\E_{b\sim\pi'_\lambda(\cdot|s)}[\ell(s,b)].
 \label{eq:path_centered_log_ratio}
\end{equation}
Since $\pi$ and $\ell$ are fixed along the path,
$\frac{d}{d\lambda}\log Z_\lambda(s)
=\sum_b\pi'_\lambda(b|s)\ell(s,b)$.
Differentiating the normalized policy therefore gives
\begin{equation}
\begin{aligned}
 \frac{d\pi'_\lambda(a|s)}{d\lambda}
 &=\pi'_\lambda(a|s)\left[
   \ell(s,a)-\frac{d}{d\lambda}\log Z_\lambda(s)\right]\\
 &=\pi'_\lambda(a|s)\left[
   \log\frac{\pi_*(a|s)}{\pi(a|s)}
   -\sum_b\pi'_\lambda(b|s)\log\frac{\pi_*(b|s)}{\pi(b|s)}
 \right]\\
 &=\pi'_\lambda(a|s)u_\lambda(s,a).
\end{aligned}
 \label{eq:path_policy_derivative}
\end{equation}
The log-ratio normalizer $Z_\lambda$ may differ from the advantage-form
normalizer by a state-dependent factor; the policy is unchanged.

\paragraph{Discounted transition operators.}
Let $P^{s'}_{sa} = P(s'|s,a)$ denote the environment transition probability of reaching
state $s'$ after taking action $a$ in state $s$.
Define the state and state--action transition kernels under $\pi'_\lambda$ by
\begin{equation}
 (P_\lambda)_s^{s'}=\sum_a \pi'_\lambda(a|s)P^{s'}_{sa},\qquad
 (\bar P_\lambda)_{sa}^{s'a'}=P^{s'}_{sa}\pi'_\lambda(a'|s').
 \label{eq:path_transition_kernels}
\end{equation}
Their discounted Green's operators are
\begin{equation}
 G_\lambda =\sum_{t\geq0}\gamma^t P_\lambda^t = (I-\gamma P_\lambda)^{-1},\qquad
 \bar G_\lambda=\sum_{t\geq0}\gamma^t\bar P_\lambda^t=(I-\gamma\bar P_\lambda)^{-1}.
 \label{eq:path_green_operators}
\end{equation}
Both series converge since $\gamma<1$, with identity matrices of the
corresponding dimensions. As a row vector, the unnormalized occupancy is
$\rho_\lambda=\rho_0G_\lambda$.
A dot denotes differentiation with respect to $\lambda$.
Differentiating $(I-\gamma P_\lambda)G_\lambda=I$ then gives
\begin{equation}
 \dot G_\lambda=\gamma G_\lambda\dot P_\lambda G_\lambda,
 \qquad
 \dot\rho_\lambda=\gamma\rho_\lambda\dot P_\lambda G_\lambda,
 \qquad
 (\dot P_\lambda)_s^{s'}
 =\sum_a \pi'_\lambda(a|s)u_\lambda(s,a)P^{s'}_{sa}.
 \label{eq:path_occupancy_derivative}
\end{equation}
To express the occupancy derivative using $\bar G_\lambda$, first expand
Eq.~\eqref{eq:path_occupancy_derivative} in components:
\begin{equation}
 \dot\rho_\lambda(s')
 =\gamma\sum_{s,a,x}\rho_\lambda(s)\pi'_\lambda(a|s)
       u_\lambda(s,a)P^x_{sa}G_\lambda(x,s').
 \label{eq:path_occupancy_components}
\end{equation}
The Green's operators are related by
\begin{equation}
 (\bar G_\lambda-I)_{sa}^{s'a'}
 =\gamma\sum_x P^x_{sa}G_\lambda(x,s')\pi'_\lambda(a'|s').
 \label{eq:path_green_kernel_connection}
\end{equation}
This follows by expanding both Green's operators in discounted
transition powers.
Multiplying Eq.~\eqref{eq:path_occupancy_components} by
$\pi'_\lambda(a'|s')$ therefore yields
\begin{equation}
 \dot\rho_\lambda(s')\pi'_\lambda(a'|s')
 =\sum_{s,a}\rho_\lambda(s)\pi'_\lambda(a|s)u_\lambda(s,a)
       (\bar G_\lambda-I)_{sa}^{s'a'}.
 \label{eq:path_occupancy_green_bridge}
\end{equation}
Summing over $a'$ recovers $\dot\rho_\lambda(s')$. Only the state
occupancy is differentiated on the left. All operators use $\pi'_\lambda$,
not the fixed base $\pi$.

\paragraph{Return derivative.}
Define the statewise improvement integrand and its action average by
\begin{align}
 B_\lambda(s,a)
 &=\widetilde A^\pi(s,a)
   -\alpha\log\frac{\pi'_\lambda(a|s)}{\pi(a|s)}
 =A_\alpha^\pi(s,a)-\alpha\log \pi'_\lambda(a|s),\\
 F_s(\lambda)&=\sum_a \pi'_\lambda(a|s)B_\lambda(s,a).
 \label{eq:path_improvement_integrand}
\end{align}
The soft performance-difference identity gives
\begin{equation}
 \widetilde\eta(\pi'_\lambda)-\widetilde\eta(\pi)
 =\sum_s\rho_\lambda(s)F_s(\lambda).
 \label{eq:path_return_difference}
\end{equation}
Since $\dot B_\lambda=-\alpha u_\lambda$ and
$\E_{\pi'_\lambda}u_\lambda=0$, differentiation yields
$\dot F_s=\E_{\pi'_\lambda}[u_\lambda B_\lambda]$.
Moreover, $\widetilde A^\pi(s,a)=\alpha\ell(s,a)+k(s)$ for an
action-independent $k(s)$, so
$B_\lambda(s,a)=\alpha(1-\lambda)\ell(s,a)+k(s)+\alpha\log Z_\lambda(s)$.
Consequently,
\begin{equation}
 \dot F_s(\lambda)
 =\alpha(1-\lambda)\Var_{a\sim \pi'_\lambda(\cdot|s)}[\ell(s,a)].
 \label{eq:path_local_derivative}
\end{equation}
Applying the product rule to Eq.~\eqref{eq:path_return_difference} gives
\begin{equation}
 \frac{d\widetilde\eta_\lambda}{d\lambda}
 =\underbrace{\sum_s\rho_\lambda(s)\dot F_s(\lambda)}_{L(\lambda)}
 +\underbrace{\sum_s\dot\rho_\lambda(s)F_s(\lambda)}_{C(\lambda)}.
 \label{eq:path_return_product_rule}
\end{equation}
Equation~\eqref{eq:path_local_derivative} gives the stated expression for
$L(\lambda)$. For $C(\lambda)$, expand
$F_{s'}(\lambda)=\sum_{a'}\pi'_\lambda(a'|s')B_\lambda(s',a')$
and substitute Eq.~\eqref{eq:path_occupancy_green_bridge} to obtain
\begin{equation}
 C(\lambda)
 =\sum_{s,a}\rho_\lambda(s)\pi'_\lambda(a|s)u_\lambda(s,a)
   \sum_{s',a'}(\bar G_\lambda-I)_{sa}^{s'a'}B_\lambda(s',a').
 \label{eq:path_occupancy_return_term}
\end{equation}
Using $B_\lambda=A_\alpha^\pi-\alpha\log\pi'_\lambda$ identifies the
second term of Eq.~\eqref{eq:path_return_derivative_decomposition},
completing the derivation.

\paragraph{Signs and the proximal limit.}
The local term $L(\lambda)$ is nonnegative on $[0,1]$. It is strictly
positive for $\lambda<1$ if $\ell(s,\cdot)$ is nonconstant at any state
with positive occupancy, and it vanishes at $\lambda=1$.
The term $C(\lambda)$ is not guaranteed to have either sign or to be small relative to the local term, so the full return need not be monotone.
At $\lambda=0$, $\pi'_0=\pi$ and
$F_s(0)=\E_{a\sim\pi}\widetilde A^\pi(s,a)=0$ for every state, so
$C(0)=0$. It follows that
\begin{equation}
 \left.\frac{d}{d\lambda}\widetilde\eta(\pi'_\lambda)\right|_{\lambda=0}
 =\frac{1}{\alpha}\sum_s\rho_\pi(s)
       \Var_{a\sim\pi(\cdot|s)}[\widetilde A^\pi(s,a)].
 \label{eq:path_proximal_return_derivative}
\end{equation}
For a fixed finite MDP with full support, these quantities are smooth:
$C(\lambda)=O(\lambda)$ near zero. If $L(0)>0$, the return therefore
increases for sufficiently small positive $\lambda$; this does
not imply monotonicity farther along the path or convergence of a practical
AAC implementation. If $\widetilde A^\pi$ vanishes everywhere, the path
is constant and both terms vanish. Appendix~\ref{app:empirical_advanced_path}
examines the empirical path shape without assuming that $C(\lambda)$ is
negligible.

\paragraph{Interpretation of the occupancy term.}
Let $H_\lambda=(\bar G_\lambda-I)B_\lambda$ denote the expected
 discounted sum of future improvement integrands conditional on the current
state and action. Since $u_\lambda$ is the centered log ratio, the
occupancy term can also be written as
\begin{equation}
 C(\lambda)=\sum_s\rho_\lambda(s)
 \operatorname{Cov}_{a\sim\pi'_\lambda(\cdot|s)}
 \bigl(\ell(s,a),H_\lambda(s,a)\bigr).
 \label{eq:path_occupancy_covariance}
\end{equation}
Transition mixing may weaken the dependence of $H_\lambda(s,a)$ on
the current action, leading to cancellations in this centered term.
This offers a possible explanation for nearly monotone paths in some
settings, not a guarantee: actions can lead to different regions of the
state space, and a long horizon can increase their differences. Moreover,
$L(1)=0$, so even a small occupancy term can determine the endpoint slope.
The following diagnostics measure path shape and endpoint regret; they do
not directly test the magnitude of $C(\lambda)$ or this mixing explanation.

\subsubsection{Empirical diagnostics}
\label{app:empirical_advanced_path}

The theory guarantees improvement over the common base policy but does not
order the family members by $\lambda$. Here we test how much return can be lost by assuming that the path is increasing. For one base policy, define the
endpoint regret
\begin{equation}
 R_{\mathrm{end}}
 =\max_{\lambda\in[0,1]}\widetilde\eta(\pi'_\lambda)
  -\widetilde\eta(\pi'_1),
 \label{eq:endpoint_regret}
\end{equation}
and normalize it by the best improvement available along the path whenever
that denominator is nonzero. This measures the loss from treating the
curve as increasing and selecting the soft-greedy endpoint, rather than merely
counting arbitrarily small negative slopes.

\paragraph{Exact random MDPs.}
We generated finite MDPs with 32 states and four actions, used
$\gamma=0.99$ and $\alpha=0.1$, and evaluated 101 equally spaced values of
$\lambda$ by solving linear systems. We considered dense random transitions, transitions
with three successors per state--action pair, and four weakly connected state
clusters. Each setting used 100 independently generated MDPs and full-support
base policies. All 300 paths satisfied
$\widetilde\eta(\pi'_\lambda)\geq\widetilde\eta(\pi)$ at every grid point, up
to numerical precision. Strict monotonicity was common but not universal:
50\%, 38\%, and 52\% of the dense, sparse, and clustered paths contained at
least one detectable decrease. The practical loss was much smaller than these
counts suggest. At random initialization, the 90th percentile of normalized
endpoint regret was respectively 0.11\%, 0.77\%, and 1.31\%.

We also repeated exact soft policy iteration,
\begin{equation}
 \pi_{k+1}=\softmax(Q_\alpha^{\pi_k}/\alpha),
\end{equation}
and reevaluated the entire path at selected iterations. The available endpoint
gain rapidly approached zero, while the exact endpoint updates monotonically
improved the training return. After the first update, the 90th percentile of
normalized endpoint regret was 0.12\%, 4.35\%, and 9.27\% for the three
transition structures. The larger value for clustered dynamics shows that
almost monotonicity is weaker when small policy changes cause substantial
occupancy redistribution. Once policy iteration converged, the entire path was
flat to numerical precision and normalized regret was no longer informative.

Table~\ref{tab:exact_path_diagnostics} summarizes the exact-MDP results.
\input{data/exact_path_diagnostics_table}

\begin{figure}[htbp]
\centering
\includegraphics[width=\textwidth]{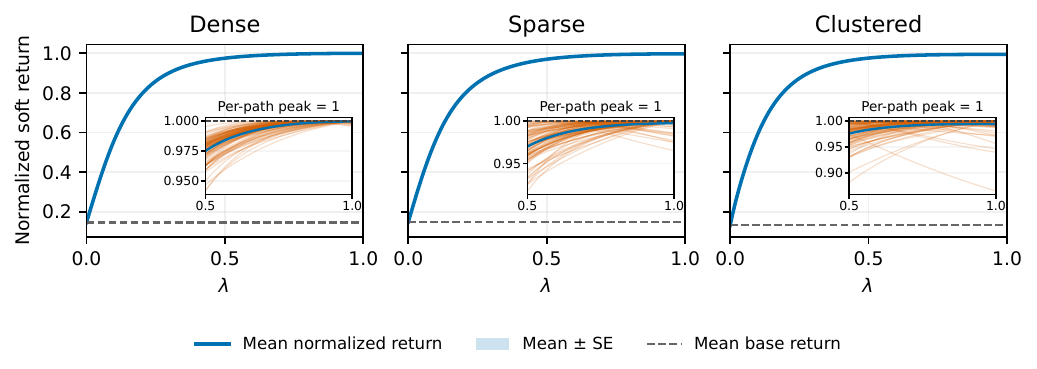}
\caption{Exact advanced-policy paths at random initialization for three
transition structures (100 MDPs each, 32 states, four actions,
$\alpha=0.1$, $\gamma=0.99$). Before averaging, each path is divided by
its own maximum return on the 101-point grid $\Lambda$:
$r_i(\lambda)=\widetilde\eta_i(\pi'_\lambda)/\max_{u\in\Lambda}\widetilde\eta_i(\pi'_u)$.
All normalization denominators are positive.
Curves show the mean normalized return at 101 values of $\lambda$;
shaded bands show one standard error across MDPs, and dashed lines mark
the mean normalized base-policy return.
Insets show all 100 normalized paths for $\lambda\in[0.5,1]$ in orange
and their means in blue, with a dashed reference at the per-path maximum, 1.
The endpoint shortfall from 1 measures loss relative to each path's grid maximum.
Inset vertical scales differ across panels. Individual peaks occur at different
values of $\lambda$, so the mean curve need not reach 1; averaging can still
hide decreases in individual paths.}
\label{fig:exact_mdp_mean_paths}
\end{figure}

Figure~\ref{fig:exact_mdp_mean_paths} shows the mean initial path for each
transition structure. Averaging can hide the small decreases in individual
paths counted in Table~\ref{tab:exact_path_diagnostics}.
The figure divides each path by its own maximum return before averaging.
The table instead divides endpoint regret by the maximum improvement over
the base policy, so its normalized regret is not the endpoint shortfall
from 1 shown in the figure.

These random MDPs need not capture the structured dependencies among
states, actions, rewards, and transitions found in control tasks or games.
Such structure may cause policy changes to redistribute state occupancy
more strongly, potentially producing larger endpoint regret even under
exact policy evaluation. The small regrets observed here therefore
should not be taken as evidence that exact advanced-policy paths are
nearly monotone in realistic environments.

\paragraph{Sampled control environment.}
We performed the same diagnostic on a saved LunarLander-v3 policy at 50k
training steps. We froze its executable policy, fitted a dedicated soft
$Q^\pi$ network for 500k transitions, and evaluated each
$\lambda\in\{0,0.05,\ldots,1\}$ on 100 paired rollout seeds. The estimated
discounted soft return increased from $13.26$ at $\lambda=0$ to $22.35$ at
$\lambda=0.2$, but then decreased to $-127.37$ at $\lambda=1$. Thus neither
almost monotonicity nor even the exact base-improvement guarantee survived this
approximate evaluation. This observation should not be read as a counterexample
to Eq.~\eqref{eq:base_improvement}: the policy path used a learned critic,
and the returns were estimated from finite rollouts. It instead illustrates
that critic and distributional error can dominate the exact statewise
improvement signal in a practical environment.

\begin{figure}[htbp]
\centering
\includegraphics[width=0.86\textwidth]{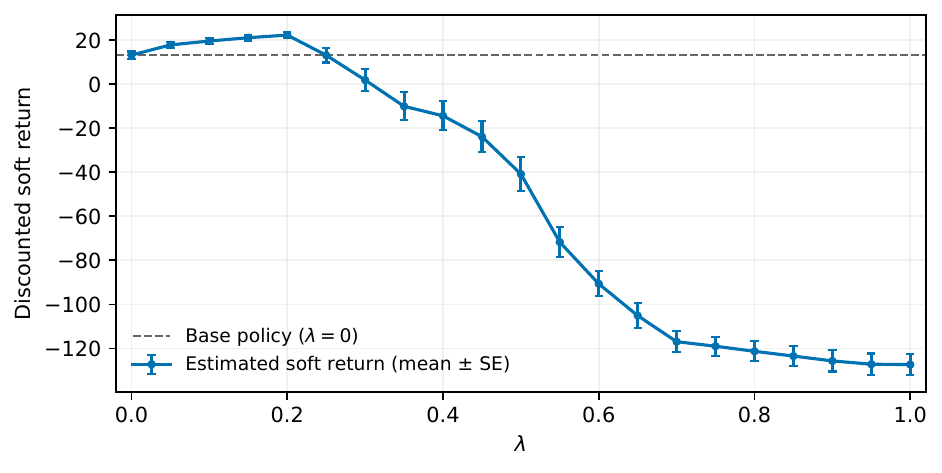}
\caption{Estimated advanced-policy soft return for a fixed LunarLander-v3
base policy saved at 50k training steps ($\alpha=0.03$, $\gamma=0.99$).
The soft critic was fitted separately for 500k transitions.
Points average 100 evaluation episodes, using the same rollout seeds across
$\lambda$; error bars show one standard error of each mean, not uncertainty
in the fitted critic. The dashed line marks the base-policy mean.
This curve uses an approximate critic and sampled returns, so it is not a
counterexample to the exact policy-improvement guarantee.}
\label{fig:lunarlander_soft_return_path}
\end{figure}

Figure~\ref{fig:lunarlander_soft_return_path} shows the estimated return
along the full evaluated path.

This example motivates limiting policy changes when the critic is
approximate, but it does not imply that AAC at large $\lambda$ or
Q-learning cannot learn. A policy farther along the plotted path may
have a lower return, while a small gradient step  may still improve 
the objective locally.
This is a possible interpretation, not an improvement guarantee for
practical AAC updates or Q-learning.

Together, these experiments show that exact paths can have small endpoint
regret even when strict monotonicity fails, while approximate values can erase
that behavior entirely. This is only a diagnostic of the exact improvement
construction. It is not used in any proof, convergence claim, or algorithmic
choice in the paper.

\subsection{Consistency and implicit-policy expressivity}
\label{app:chasing_expressivity}

AAC treats $(\pi_\theta,Q_\phi)$ as independent coordinates of the
composed policy.  For fixed function classes, write
\begin{align}
 \mathcal P_{\mathrm{free}}
 &=\left\{\pi'_{\lambda}(\pi_\theta,Q_\phi):
       \theta\in\Theta,\phi\in\Phi\right\},\\
 \mathcal P_{\mathrm{cons}}
 &=\left\{\pi'_{\lambda}(\pi_\theta,Q^{\pi_\theta}):
       \theta\in\Theta\right\},
\end{align}
where the second expression is interpreted with an exact evaluator or with a
critic class capable of representing it. Evaluation-consistent pairs form a
constrained subset of the free coordinate pairs. When the chosen function classes can represent these consistent pairs, this
implies $\mathcal P_{\mathrm{cons}}\subseteq\mathcal P_{\mathrm{free}}$;
strict inclusion depends on the MDP and parameterization and is not assumed.
Chasing moves toward this constrained set rather than imposing exact
consistency at every training step. We test whether that movement can
change the modality of the represented advanced policy before its executable
projection.

\paragraph{Exact diagnostic MDP.}
The environment has ten finite states and a scalar continuous action
$a\in[-4,4]$.  The action controls clockwise versus counterclockwise movement
on a state ring through a sigmoid transition probability.  Rewards are
constructed from state-dependent mixtures of Gaussian action modes.  We use
both two-mode and three-mode rewards.  The base and fast actors are
state-dependent single Gaussians, while $Q_\phi$ uses a smooth radial-basis
representation.  Action expectations are deterministic sums on a 301-point
grid. Policy values and evaluation Q-functions are computed by solving
the Bellman linear system for this discretized model, avoiding rollout
noise and learned-critic evaluation error within that model. Continuous
action integrals are still approximated by the grid. The independent base
coordinate $Q_\phi$ is learned through its radial-basis coefficients and is
not itself an exact policy evaluation. Return differences reported in this
diagnostic use the normalized soft return $(1-\gamma)\widetilde\eta$.

For every environment and initialization, $\tau=0$ provides the paired AAC
comparison. Positive $\tau$ moves the base Gaussian and base Q toward the fast
Gaussian and its exact evaluation critic.  We use 30 paired initializations.
At each state we count significant local maxima of the exact normalized
$\pi'(a| s)$.  A peak must have at least $3\%$ of the largest density,
$2\%$ relative prominence, and five action-grid points of separation.  These
thresholds exclude small numerical fluctuations where the density is very low.

\begin{table}[h]
\centering
\caption{Modality of the implicit advanced policy in the two-mode MDP
($\lambda=0.7$, strong action-dependent transitions).  ``Multimodal states''
is averaged over 30 initializations and ten states.}
\label{tab:bimodal_chasing_modality}
\begin{tabular}{lrrr}
\toprule
Setting & Modes per state & Multimodal states & Runs with any multimodality\\
\midrule
AAC, $\tau=0$       & 1.980 & 96.7\% & 30/30\\
Chasing, $\tau=0.001$  & 1.447 & 44.7\% & 20/30\\
Chasing, $\tau=0.01$   & 1.030 &  3.0\% &  9/30\\
\bottomrule
\end{tabular}
\end{table}

Table~\ref{tab:bimodal_chasing_modality} shows a large one-directional change.
For $\tau=0.01$, 290 of the 300 state policies were multimodal under AAC.
Under chasing, 281 of those became unimodal and only nine remained multimodal;
no state that was unimodal under AAC became multimodal under chasing. 
Every one of the 30 paired runs had a
lower mean mode count under chasing.  The corresponding mean change in fast
policy return was only $-2.45\times10^{-4}$, whereas the RMSE between the base
Q and the exact fast-policy Q decreased by $3.44$. Thus stronger chasing
reduced both this Q mismatch and the number of modes, while executable
returns remained close under the shared single-Gaussian projection.
These measurements do not isolate reduced Q mismatch as the cause of the
modality change, or establish a return benefit from the extra modes.

\begin{figure}[htbp]
\centering
\includegraphics[width=\textwidth]{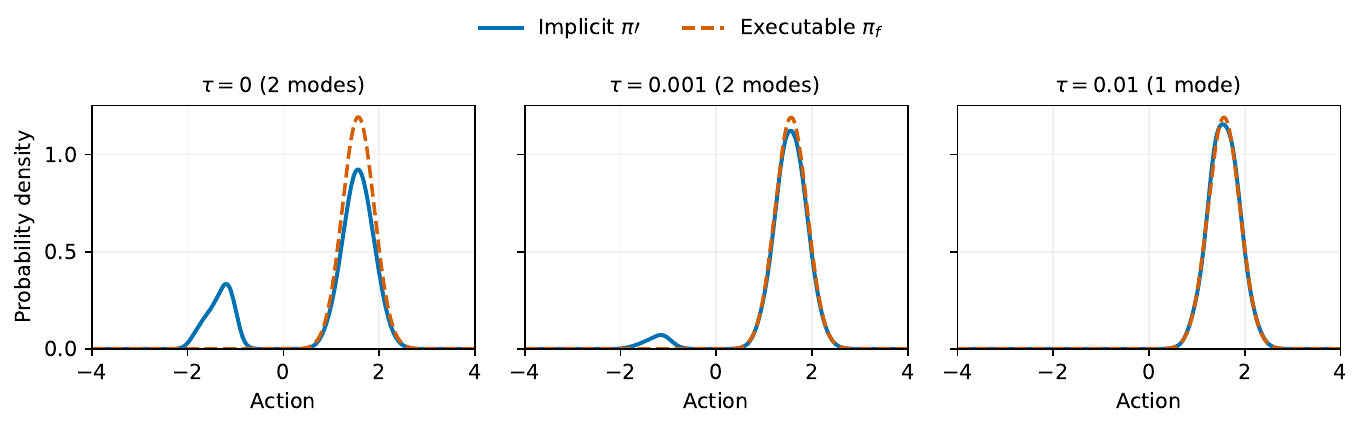}
\caption{Illustrative policy densities in the 10-state bimodal diagnostic
($\lambda=0.7$, seed 19, state 0, after 600 updates).
Solid curves show the implicit advanced policy $\pi'$; dashed curves show
its single-Gaussian executable approximation $\pi_f$.
The same initialization and state are used across chasing strengths.
The seed minimizes the squared distance of its three multimodal-state
fractions from their ensemble means; the state is selected in the same way
using mode counts within that seed, with ties broken by index.
Densities are normalized on the action grid. This paired example
illustrates the distributional change, rather than a return advantage.}
\label{fig:modality_distribution_example}
\end{figure}

The three-mode experiment gives the same direction with a smaller effect.  At
$\lambda=0.3$ and medium transition coupling, AAC represented 2.77 modes
per state and was multimodal at every state.  Strong chasing represented 2.52
modes per state and was multimodal at 76\% of states.  Seventy-two state
policies changed from multimodal to unimodal and none changed in the reverse
direction.  The mean fast-return change was $-6.04\times10^{-4}$.

The supplementary folder \emph{diagnostics/modality/} contains the
original per-run results, settings, and instructions for rerunning this
diagnostic and recomputing its summary tables.

\paragraph{Scope of the result.}
This diagnostic supports a qualitative claim: relaxing policy--critic
consistency can preserve implicit advanced policies with different and richer
modality, even when both variants use the same executable Gaussian class.  It
does not prove that AAC is generally better than chasing, nor does it by
itself prove strict set inclusion: a different optimizer or initialization
might reach another policy in the consistent policy class.  A broader sweep likewise
found no systematic return ordering.  The appropriate interpretation is an
expressivity--consistency trade-off whose difference in the implicit policies can largely disappear
when both are fitted by a restricted fast-policy class.

\subsection{Additional control Diagnostics}
\label{app:preliminary_control}

These supplementary diagnostics provide additional detail on learning behavior 
across environments and settings. Budgets, training-seed counts, and evaluation
protocols vary across these diagnostics and are stated with the results;
they should not be treated as a matched-budget comparison. The four-seed
Pendulum diagnostic reports population standard deviations, whereas the
Acrobot sweep and additional result tables report sample standard deviations.

\paragraph{Discrete control.}
We evaluated Acrobot-v1, MountainCar-v0, and LunarLander-v3 using MLPs with two
hidden layers, replay, exact action sums, and separate Adam optimizers for the
actor and Q channels. Table~\ref{tab:acrobot_full_sweep} gives the current
five-seed Acrobot sweep. On MountainCar,
AAC with $\alpha=10^{-3}$ obtained approximately $-102.6$, $-109.8$, and
$-97.4$ at $\lambda\in\{0.3,0.7,1\}$, averaged over three seeds. Larger
temperatures were less reliable. Additional LunarLander runs reached returns
above $200$ for endpoint and intermediate settings.

\begin{table*}[t]
\centering
\caption{Acrobot-v1 final advanced-policy return after 50k environment steps
(mean and sample standard deviation over five training seeds; higher is
better). All methods use $\alpha=10^{-3}$, $\gamma=.99$, exact discrete action
sums, $q_{\mathrm{pg\text{-}uniform\text{-}mix}}=.1$, and an advanced-Q
bootstrap. Direct AAC uses $\tau=0$ and direct learning rates $3\times10^{-4}$.
Chasing AAC uses $\tau=.01$ with the direct rates disabled. Combined AAC uses
$\tau=.01$ and direct learning rates $10^{-5}$.}
\label{tab:acrobot_full_sweep}
\begin{tabular}{c r r r}
\toprule
$\lambda$ & Direct AAC & Chasing AAC & Combined AAC\\
\midrule
0.0 & $-79.9\pm4.7$ & $-484.7\pm32.4$ & $-84.1\pm9.0$\\
0.1 & $-109.3\pm31.2$ & $-81.8\pm7.5$ & $-88.9\pm8.0$\\
0.2 & $-89.9\pm19.4$ & $-85.5\pm5.0$ & $-84.1\pm3.2$\\
0.3 & $-81.3\pm2.0$ & $-83.0\pm8.2$ & $-86.1\pm10.9$\\
0.4 & $-81.7\pm9.7$ & $-85.7\pm5.0$ & $-80.9\pm4.6$\\
0.5 & $-77.6\pm5.8$ & $-101.7\pm20.3$ & $-88.3\pm7.3$\\
0.6 & $-81.7\pm2.5$ & $-86.6\pm7.8$ & $-79.4\pm3.4$\\
0.7 & $-78.4\pm4.8$ & $-88.3\pm5.5$ & $-81.5\pm4.9$\\
0.8 & $-87.9\pm13.8$ & $-85.1\pm6.5$ & $-83.2\pm5.3$\\
0.9 & $-86.8\pm20.6$ & $-83.6\pm7.7$ & $-85.4\pm3.9$\\
1.0 & $-78.3\pm2.6$ & $-81.9\pm6.0$ & $-80.7\pm5.9$\\
\bottomrule
\end{tabular}
\end{table*}

\paragraph{Base and fast policies on Pendulum.}
We trained Pendulum-v1 for 50k environment steps using squashed-Gaussian actors
and Eq.~\eqref{eq:reverse_kl}. Table~\ref{tab:pendulum_preliminary} reports
four training seeds under deterministic evaluation on ten fixed initial states
per seed. All successful executable policies converged to nearly the same
return. The important diagnostic is instead the separation between the base
and fast policies under AAC without chasing: the composed executable policy can be
good even when the base coordinate is not.

\begin{table}[h]
\centering
\caption{Pendulum-v1 returns after 50k steps (mean and population standard
deviation over four training seeds). Higher is better. ``Fast'' denotes the
executable approximation to the advanced policy.}
\label{tab:pendulum_preliminary}
\begin{tabular}{llrr}
\toprule
Method & $\lambda$ & Base $\pi$ & Fast $\pi_{\theta_f}$\\
\midrule
AAC & 0.0 & $-179.6\pm9.1$ & $-179.8\pm9.5$\\
        & 0.3 & $-316.0\pm161.0$ & $-179.8\pm9.3$\\
        & 0.7 & $-998.5\pm485.6$ & $-180.1\pm9.7$\\
        & 1.0 & $-1440.6\pm134.7$ & $-180.6\pm9.8$\\
\midrule
Chasing & 0.0 & $-1371.5\pm52.2$ & $-1366.8\pm59.3$\\
             & 0.3 & $-180.2\pm9.6$ & $-180.1\pm9.3$\\
             & 0.7 & $-179.8\pm9.5$ & $-180.1\pm9.4$\\
             & 1.0 & $-180.0\pm9.1$ & $-180.4\pm9.2$\\
\midrule
Combined & 0.0 & $-180.7\pm10.4$ & $-180.2\pm10.2$\\
         & 0.3 & $-180.3\pm9.1$ & $-180.2\pm9.6$\\
         & 0.7 & $-181.0\pm8.3$ & $-180.3\pm8.2$\\
         & 1.0 & $-179.9\pm8.9$ & $-179.4\pm8.3$\\
\bottomrule
\end{tabular}
\end{table}

\input{sections/appendix_bookkeeping}

\input{sections/appendix_walker2d}

\input{sections/appendix_published_baselines}

%% file: data/exact_path_diagnostics_table.tex
\begin{table}[htbp]
\centering
\small
\caption{Exact random-MDP path diagnostics (100 MDPs per transition
structure, 32 states, four actions, $\alpha=0.1$, $\gamma=0.99$).
The second column counts paths with a detectable decrease at initialization.
The last two columns give the 90th percentile of endpoint regret divided by
the largest improvement over the base policy on the evaluated path,
at initialization and after one exact soft policy-iteration update.
All quantities use the 101-point $\lambda$ grid.}
\label{tab:exact_path_diagnostics}
\begin{tabular}{lrrr}
\toprule
& Paths with & \multicolumn{2}{c}{Normalized endpoint regret (90th percentile)}\\
Transition structure & a decrease & Initial & After one update\\
\midrule
Dense & 50\% & 0.11\% & 0.12\% \\
Sparse & 38\% & 0.77\% & 4.35\% \\
Clustered & 52\% & 1.31\% & 9.27\% \\
\bottomrule
\end{tabular}
\end{table}

%% file: sections/appendix_bookkeeping.tex
\subsubsection{Additional result tables: CartPole, HalfCheetah, MountainCar, and Pendulum}
\label{app:bookkeeping_control}

The following tables record the available control results and the runs they come from.  Parentheses give the number of training seeds, and the
reported uncertainty is the sample standard deviation.  They should not be read as a
cross-environment performance ranking.

\begin{table*}[t]
\centering
\caption{CartPole-v1 advanced-policy return after 50k environment steps.
All entries use $\alpha=10^{-3}$, exact action sums,
$q_{\rm pg\text{-}uniform\text{-}mix}=.1$, target-Q bootstrap, and no warm-up.
Direct and Combined are the new five-seed runs; Chasing combines its matching
three-seed sweep with the matching seed-4/5 extension.  Chasing uses $\tau=.01$;
Combined uses $\tau=.01$ and direct learning rates $10^{-5}$.}
\label{tab:cartpole_bookkeeping}
\begin{tabular}{c r r r}
\toprule
$\lambda$ & Direct AAC & Chasing AAC & Combined AAC\\
\midrule
0.0 & $382.8\pm105.8$ (5) & $25.2\pm5.6$ (5) & $340.9\pm152.4$ (5)\\
0.1 & $325.2\pm125.9$ (5) & $500.0\pm0.0$ (5) & $408.5\pm74.4$ (5)\\
0.2 & $377.0\pm168.8$ (5) & $387.1\pm112.8$ (5) & $470.9\pm59.4$ (5)\\
0.3 & $464.6\pm79.1$ (5) & $490.5\pm21.2$ (5) & $401.0\pm135.6$ (5)\\
0.4 & $379.8\pm115.0$ (5) & $440.5\pm81.1$ (5) & $414.6\pm136.6$ (5)\\
0.5 & $373.0\pm179.2$ (5) & $412.8\pm77.5$ (5) & $425.1\pm122.8$ (5)\\
0.6 & $500.0\pm0.0$ (5) & $443.5\pm114.3$ (5) & $405.0\pm119.3$ (5)\\
0.7 & $313.1\pm158.1$ (5) & $305.6\pm181.2$ (5) & $333.8\pm126.3$ (5)\\
0.8 & $438.1\pm138.4$ (5) & $372.8\pm132.2$ (5) & $381.4\pm115.5$ (5)\\
0.9 & $437.6\pm87.2$ (5) & $356.3\pm148.2$ (5) & $351.1\pm181.2$ (5)\\
1.0 & $356.2\pm137.8$ (5) & $465.0\pm78.3$ (5) & $382.3\pm169.0$ (5)\\
\bottomrule
\end{tabular}
\end{table*}

\begin{table*}[t]
\centering
\caption{HalfCheetah-v5 final advanced-policy return after 200k environment
steps. Entries are the mean and sample standard deviation over five training
seeds under deterministic evaluation with ten episodes per seed. All methods
use $\alpha=.2$, two 256-unit hidden layers, target-Q bootstrap, and behavior
sampling from the advanced policy. Direct uses learning rates $3\times10^{-4}$
and $\tau=0$; Chasing uses $\tau=.01$ with Direct disabled; Combined uses
$\tau=.001$ and Direct learning rates $3\times10^{-4}$. The reported runs use 
matching recorded algorithm and training settings and are aggregated across 
five independent training seeds.}
\label{tab:halfcheetah_bookkeeping}
\begin{tabular}{c r r r}
\toprule
$\lambda$ & Direct AAC & Chasing AAC & Combined AAC\\
\midrule
0.0 & $6179\pm900$ & $-4\pm11$ & $5944\pm1170$\\
0.1 & $5673\pm702$ & $5926\pm1037$ & $5868\pm839$\\
0.2 & $5967\pm1158$ & $6010\pm838$ & $6274\pm942$\\
0.3 & $6175\pm831$ & $6231\pm771$ & $6443\pm663$\\
0.4 & $5963\pm1097$ & $6750\pm970$ & $5945\pm899$\\
0.5 & $5748\pm847$ & $6492\pm951$ & $6380\pm1202$\\
0.6 & $5891\pm806$ & $6411\pm897$ & $6141\pm904$\\
0.7 & $5997\pm1255$ & $6283\pm773$ & $6432\pm1009$\\
0.8 & $5914\pm525$ & $6508\pm1154$ & $6215\pm525$\\
0.9 & $5986\pm621$ & $6066\pm651$ & $5762\pm1174$\\
1.0 & $6168\pm1039$ & $6745\pm932$ & $6035\pm870$\\
\bottomrule
\end{tabular}
\end{table*}

\begin{table*}[t]
\centering
\caption{MountainCar-v0 advanced-policy return after 500k environment steps.
Entries are the mean and sample standard deviation over five training seeds,
with stochastic evaluation over ten episodes per seed.  All methods use
$\alpha=10^{-3}$, exact action sums, $q_{\rm pg\text{-}uniform\text{-}mix}=.1$,
target-Q bootstrap, and no warm-up.  Direct uses learning rates $3\times10^{-4}$
and $\tau=0$; Chasing uses $\tau=.01$ with Direct disabled; Combined uses
$\tau=.01$ and Direct learning rates $10^{-5}$.}
\label{tab:mountaincar_bookkeeping}
\begin{tabular}{c r r r}
\toprule
$\lambda$ & Direct AAC & Chasing AAC & Combined AAC\\
\midrule
0.0 & $-118.7\pm45.5$ (5) & $-200.0\pm0.0$ (5) & $-200.0\pm0.0$ (5)\\
0.1 & $-156.9\pm40.3$ (5) & $-117.8\pm46.0$ (5) & $-98.2\pm2.4$ (5)\\
0.2 & $-117.9\pm26.7$ (5) & $-118.8\pm45.4$ (5) & $-118.6\pm45.5$ (5)\\
0.3 & $-119.5\pm45.1$ (5) & $-106.6\pm15.8$ (5) & $-138.5\pm56.1$ (5)\\
0.4 & $-118.8\pm45.4$ (5) & $-138.8\pm55.9$ (5) & $-159.4\pm55.7$ (5)\\
0.5 & $-119.7\pm44.9$ (5) & $-128.9\pm43.9$ (5) & $-101.0\pm4.8$ (5)\\
0.6 & $-118.6\pm45.5$ (5) & $-117.6\pm46.1$ (5) & $-121.0\pm44.3$ (5)\\
0.7 & $-99.2\pm2.3$ (5) & $-140.4\pm54.6$ (5) & $-140.5\pm54.3$ (5)\\
0.8 & $-140.2\pm54.8$ (5) & $-118.1\pm45.8$ (5) & $-102.0\pm5.7$ (5)\\
0.9 & $-99.3\pm3.0$ (5) & $-99.3\pm1.0$ (5) & $-99.7\pm2.8$ (5)\\
1.0 & $-98.8\pm1.2$ (5) & $-120.6\pm44.7$ (5) & $-138.6\pm56.0$ (5)\\
\bottomrule
\end{tabular}
\end{table*}

\begin{table*}[t]
\centering
\caption{Pendulum-v1 executable advanced-policy return after 50k environment
steps.  Results are aggregated over the training seeds available for each 
reported condition. The $\lambda=0.5$ entries include the completed seed-1 
runs for all three
methods; other rows retain the stated seed counts from those records.}
\label{tab:pendulum_bookkeeping}
\begin{tabular}{c r r r}
\toprule
$\lambda$ & Direct AAC & Chasing AAC & Combined AAC\\
\midrule
0.0 & $-176.9\pm11.5$ (5) & $-1367.3\pm59.3$ (5) & $-177.5\pm11.9$ (5)\\
0.1 & $-177.1\pm11.0$ (5) & $-180.6\pm12.0$ (4) & $-180.4\pm11.1$ (4)\\
0.2 & $-177.1\pm10.8$ (5) & $-180.2\pm11.9$ (4) & $-180.2\pm9.7$ (4)\\
0.3 & $-177.2\pm11.0$ (5) & $-177.4\pm11.1$ (5) & $-177.3\pm11.6$ (5)\\
0.4 & $-177.7\pm11.5$ (5) & $-179.7\pm10.2$ (4) & $-179.6\pm11.4$ (4)\\
0.5 & $-178.2\pm11.5$ (5) & $-177.5\pm10.8$ (5) & $-177.8\pm11.8$ (5)\\
0.6 & $-177.3\pm11.0$ (5) & $-179.2\pm11.2$ (4) & $-180.3\pm10.9$ (4)\\
0.7 & $-177.4\pm11.4$ (5) & $-177.2\pm11.4$ (5) & $-177.5\pm10.3$ (5)\\
0.8 & $-177.7\pm11.4$ (5) & $-179.7\pm11.5$ (4) & $-180.6\pm11.2$ (4)\\
0.9 & $-178.0\pm10.9$ (5) & $-180.4\pm10.0$ (4) & $-179.8\pm11.2$ (4)\\
1.0 & $-177.8\pm11.7$ (5) & $-177.8\pm10.9$ (5) & $-176.8\pm10.1$ (5)\\
\bottomrule
\end{tabular}
\end{table*}

\paragraph{Completed results at the representative value.}
The CartPole and Pendulum entries at $\lambda=0.5$ in
Table~\ref{tab:main_control_summary} each use seeds 1--5. Duplicate copies 
of the same run are counted once. Within each reported condition, 
the recorded algorithm and training settings agree. Available run 
configurations and evaluation records were checked before aggregation. 
CartPole uses stochastic evaluation, whereas Pendulum uses
deterministic evaluation, both over ten episodes per seed. Both use hidden
width 256, learning starts at 1000 steps, and no separate PG warm-up.
Pendulum uses $\alpha=0.2$, target-Q bootstrap, Direct rates
$3\times10^{-4}$ with $\tau=0$, Chasing with Direct disabled and
$\tau=0.01$, and Combined rates $3\times10^{-4}$ with $\tau=0.001$.

\paragraph{LunarLander at $\lambda=0.5$.}
At 500k steps, the available LunarLander-v3 results use seeds 1--5 for
Direct and Chasing, and seeds 1--3 for Combined. They are reported with
these unequal counts rather than supplemented with incompatible runs.
All use hidden width 256, $\alpha=10^{-3}$, $\gamma=0.99$,
$q_{\rm pg\text{-}uniform\text{-}mix}=0.1$, target-Q bootstrap, learning
starts at 1000 steps, and stochastic ten-episode evaluation. Direct uses
rates $3\times10^{-4}$ and $\tau=0$; Combined uses the same direct rates
and $\tau=0.001$. Both retain a 10k-step PG warm-up. Chasing disables
Direct and uses $\tau=0.01$ without PG warm-up. Thus the comparison
includes method-specific update schedules as well as different seed counts.

\begin{table}[t]
\centering
\caption{LunarLander-v3 final evaluation at $\lambda=0.5$ after 500k steps.
Each seed entry is a ten-episode mean. Aggregate uncertainty is the sample
standard deviation across seed means. Missing seed results are not estimated.}
\label{tab:lunarlander_representative}
\begin{tabular}{lrrr}
\toprule
Seed & Direct AAC & Chasing AAC & Combined AAC\\
\midrule
1 & 282.8 & 280.9 & 1.8\\
2 & 237.1 & 279.2 & 218.4\\
3 & 269.0 & 290.3 & 284.0\\
4 & 196.1 & 283.2 & ---\\
5 & 197.5 & 218.4 & ---\\
\midrule
Mean $\pm$ SD & $236.5\pm39.9$ & $270.4\pm29.4$ & $168.0\pm147.7$\\
Seeds & 5 & 5 & 3\\
\bottomrule
\end{tabular}
\end{table}

%% file: sections/appendix_walker2d.tex
\subsection{Walker2d: one-million-step comparison}
\label{app:walker2d_results}

\paragraph{Protocol and aggregation.}
We report 80 completed Walker2d-v5 runs: Direct at
$\lambda\in\{0,0.1,0.3,0.5,0.7,1\}$, and Chasing and Combined at
$\lambda\in\{0.1,0.3,0.5,0.7,1\}$, with training seeds 21--25 for every
reported condition. The Combined $\lambda=0$ condition is excluded from
this comparison; Chasing at $\lambda=0$ was not run. Each run uses 1M
environment steps, and the reported score is the final checkpoint's
undiscounted environment return, without the entropy bonus, averaged over
ten deterministic episodes of the executable fast policy $\pi_f$.
We report the arithmetic mean and sample standard deviation of the five
seed means, rather than pooling the 50 episodes or selecting the best
checkpoint. All five seeds of each reported condition are retained.

The base and fast networks have two hidden layers of width 256. Shared
settings are $\gamma=0.99$, fixed $\alpha=0.2$, batch size 256, replay
capacity 1M, and uniform exploration probability 0.005. Learning starts
after 10k steps, with no initial random-action phase or separate PG warm-up.
There is one gradient update per environment step thereafter. Both fast
learning rates are $3\times10^{-4}$; target-network Polyak updates use
0.005. Soft TD uses the target fast actor and critic, and bootstraps
through time-limit truncations. Evaluation occurs every 100k steps.
Direct uses actor/Q learning rates $3\times10^{-4}$ and $\tau=0$;
Chasing disables those updates and uses $\tau=0.01$; Combined uses
actor/Q learning rates $10^{-5}$ and $\tau=0.01$. The channel learning-rate
scaling described in Section~\ref{sec:algorithm} applies. No explicit
Q gauge regularizer is used. Each worker uses one Torch CPU thread.

\paragraph{Included runs and checks.}
These settings were fixed after preliminary screening, rather than a
search over all settings. Earlier screening runs with different seeds or settings
are excluded from this comparison. Recorded algorithm and training configurations 
agree across runs except for the intended method and \(\lambda\) differences. 
Run settings, evaluation logs, and final results were checked before computing 
the reported statistics.

\begin{table}[t]
\centering
\caption{Walker2d-v5 executable-policy return after 1M environment steps.
Entries are mean $\pm$ sample standard deviation over five training seeds,
with ten deterministic evaluation episodes per seed. Only Direct is reported
at $\lambda=0$. All five seeds are retained within every reported condition.}
\label{tab:walker2d_basic_sweep}
\begin{tabular}{crrr}
\toprule
$\lambda$ & Direct AAC & Chasing AAC & Combined AAC\\
\midrule
0.0 & $3605\pm1224$ & --- & ---\\
0.1 & $2129\pm823$ & $2137\pm1457$ & $3041\pm541$\\
0.3 & $2458\pm781$ & $3389\pm1238$ & $2306\pm1831$\\
0.5 & $3422\pm1436$ & $2778\pm560$ & $3290\pm924$\\
0.7 & $2099\pm968$ & $2276\pm1377$ & $2394\pm1042$\\
1.0 & $3627\pm638$ & $2465\pm1572$ & $3299\pm1831$\\
\bottomrule
\end{tabular}
\end{table}

\paragraph{Results and seed variation.}
Table~\ref{tab:walker2d_basic_sweep} gives mean scores and standard deviations, and
Table~\ref{tab:walker2d_seed_scores} gives all individual seed means.
Direct's $\lambda=1$ scores range from 2672 to 4142, whereas the corresponding
Chasing and Combined settings each contain one near-zero final score.
At $\lambda=0.5$, Chasing has a narrower spread, while Direct and Combined
have larger means. Low-scoring seeds are included in every reported mean
and uncertainty estimate. These observations describe the tested settings;
they do not imply a general ordering of the algorithms.
Combined has the highest mean at $\lambda=0.1$ (3041) and $0.7$ (2394),
Chasing at $\lambda=0.3$ (3389), and Direct at $\lambda=0.5$ and $1$.
At $\lambda=0.3$, Combined includes scores of 665 and 6, giving a large
across-seed spread despite the other three seeds exceeding 3200. At $\lambda=1$, four Combined
seeds score between 3881 and 4268 and one scores 34.
The one-million-step budget follows a short pilot that did not approach
a stable plateau; it is not a claim that all conditions have converged.

\begin{table}[t]
\centering
\small
\caption{Walker2d-v5 final executable-policy scores for every training seed.
Each entry averages ten deterministic episodes and is rounded to the nearest
integer; aggregate statistics use unrounded values.}
\label{tab:walker2d_seed_scores}
\begin{tabular}{lrrrrrr}
\toprule
Method & $\lambda$ & Seed 21 & Seed 22 & Seed 23 & Seed 24 & Seed 25\\
\midrule
Direct & 0.0 & 4567 & 5025 & 2484 & 3682 & 2267\\
Direct & 0.1 & 2034 & 1989 & 1464 & 3537 & 1623\\
Direct & 0.3 & 3225 & 1139 & 2562 & 2721 & 2643\\
Direct & 0.5 & 3736 & 4114 & 4607 & 3717 & 936\\
Direct & 0.7 & 3226 & 2414 & 1940 & 2334 & 583\\
Direct & 1.0 & 3271 & 4142 & 3971 & 4079 & 2672\\
\midrule
Chasing & 0.1 & 1715 & -3 & 2404 & 4004 & 2564\\
Chasing & 0.3 & 3902 & 1599 & 3973 & 4755 & 2717\\
Chasing & 0.5 & 3470 & 2377 & 2384 & 2354 & 3305\\
Chasing & 0.7 & 2546 & 152 & 3649 & 3225 & 1810\\
Chasing & 1.0 & 2900 & 2385 & 2650 & 4376 & 14\\
\midrule
Combined & 0.1 & 3659 & 2877 & 3132 & 3318 & 2217\\
Combined & 0.3 & 3223 & 3872 & 665 & 6 & 3767\\
Combined & 0.5 & 2888 & 4814 & 2351 & 3072 & 3326\\
Combined & 0.7 & 945 & 3265 & 2138 & 2077 & 3543\\
Combined & 1.0 & 4111 & 4203 & 3881 & 34 & 4268\\
\bottomrule
\end{tabular}
\end{table}

\begin{figure}[t]
\centering
\includegraphics[width=\textwidth]{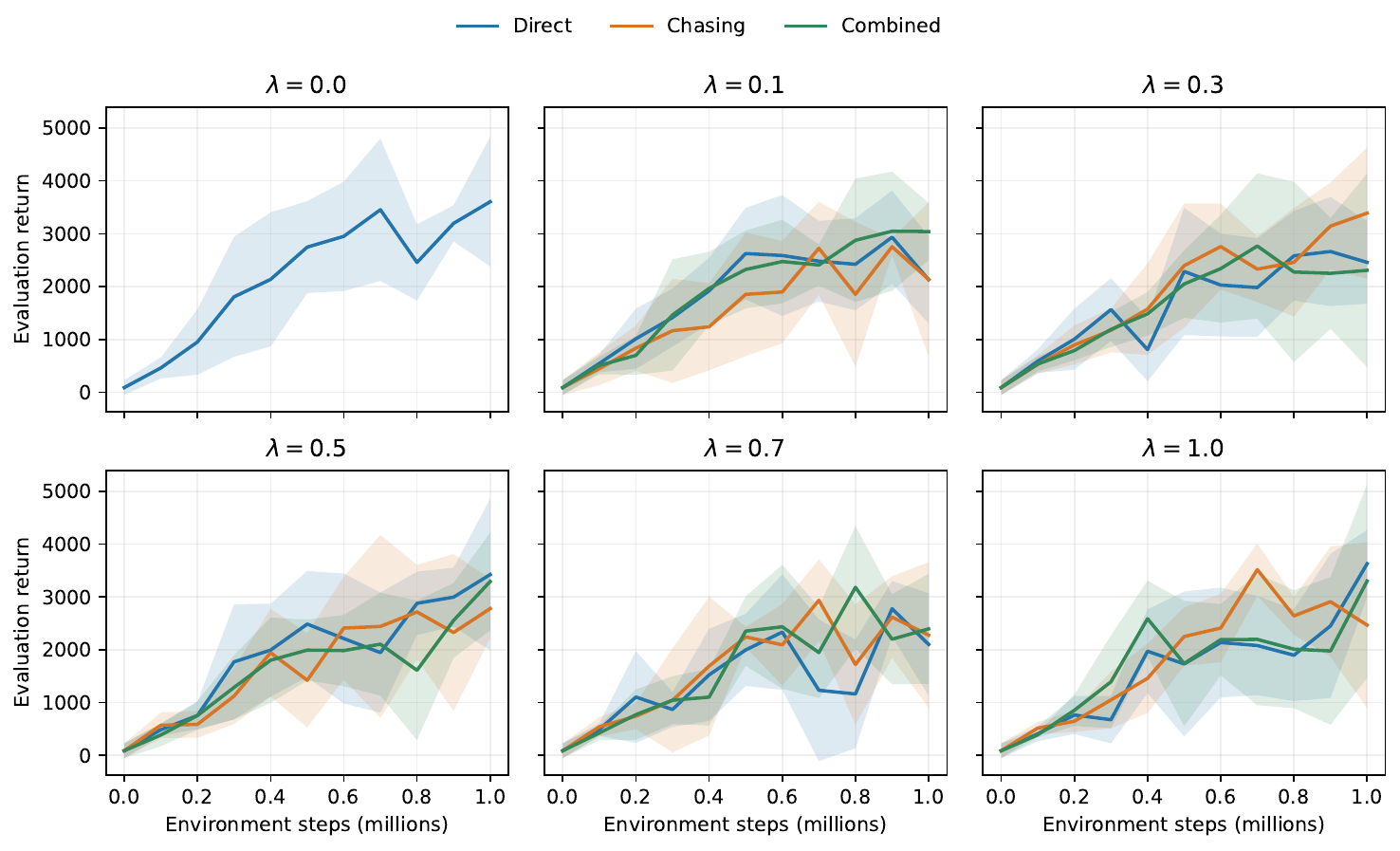}
\caption{Walker2d-v5 executable-policy learning curves. Lines show the mean
of five training-seed evaluation means; bands show one sample standard
deviation across seeds, not a confidence interval. Each evaluation uses ten
deterministic episodes. Only Direct is reported at $\lambda=0$.}
\label{fig:walker2d_basic_learning_curves}
\end{figure}

%% file: sections/appendix_published_baselines.tex
\clearpage
\subsection{Published baseline results for context}
\label{app:published_baselines}

Table~\ref{tab:published_baselines} collects reference scores from two
studies that evaluate several algorithms within a common setup.
\citet[Table~1]{Wang2019MBRLBenchmark} report HalfCheetah results at 200k
steps, matching our interaction budget. Their results use four seeds and
a 5k-step averaging window. \citet[Table~1]{Fujimoto2023SALE} report SAC
and TD3 results on unmodified Gym v4 environments, including Walker2d at
our 1M-step budget. They evaluate deterministic policies every 5k steps
for 10 episodes and average over 10 seeds.

We also list our $\lambda=0.5$ results from
Table~\ref{tab:main_control_summary}, keeping the same setting for all three
AAC modes rather than averaging over $\lambda$ or selecting the best value.

\begin{table}[htbp]
\centering
\small
\caption{Published baseline scores and AAC results for context, with
unmatched environments and evaluation protocols. Wang et al. report
means and standard deviations over four seeds; Fujimoto et al. report
means and approximate 95\% confidence intervals over ten seeds
($1.96$ times the standard error). AAC entries reproduce
Table~\ref{tab:main_control_summary} at $\lambda=0.5$, with means and sample
standard deviations over five seeds.}
\label{tab:published_baselines}
\begin{tabular}{llllr}
\toprule
Source & Environment & Steps & Method & Return\\
\midrule
\citet{Wang2019MBRLBenchmark} & HalfCheetah & 200k & TD3 & $3614.3\pm82.1$\\
 & HalfCheetah & 200k & SAC & $4000.7\pm202.1$\\
 & HalfCheetah & 200k & PPO & $17.2\pm84.4$\\
\midrule
\citet{Fujimoto2023SALE} & HalfCheetah-v4 & 300k & TD3 & $7715\pm633$\\
 & HalfCheetah-v4 & 300k & SAC & $8052\pm515$\\
 & HalfCheetah-v4 & 1M & TD3 & $10574\pm897$\\
 & HalfCheetah-v4 & 1M & SAC & $10484\pm659$\\
 & Walker2d-v4 & 1M & TD3 & $3946\pm292$\\
 & Walker2d-v4 & 1M & SAC & $4314\pm256$\\
\midrule
This work & HalfCheetah-v5 & 200k & Direct AAC & $5748\pm847$\\
($\lambda=0.5$) & HalfCheetah-v5 & 200k & Chasing AAC & $6492\pm951$\\
 & HalfCheetah-v5 & 200k & Combined AAC & $6380\pm1202$\\
 & Walker2d-v5 & 1M & Direct AAC & $3422\pm1436$\\
 & Walker2d-v5 & 1M & Chasing AAC & $2778\pm560$\\
 & Walker2d-v5 & 1M & Combined AAC & $3290\pm924$\\
\bottomrule
\end{tabular}
\end{table}

Our experiments use Gymnasium v5 environments and final-checkpoint
scores, so these results do not provide a controlled comparison with AAC.
The HalfCheetah results from \citet{Fujimoto2023SALE} also use longer
training budgets. The low PPO score at 200k steps describes that study's
setting, not PPO performance in general. We omit the Walker2d results from
\citet{Wang2019MBRLBenchmark}, whose benchmark changes the reward function
and separately tests early termination.